\documentclass[11pt]{article}
\usepackage[utf8]{inputenc}
\usepackage{graphicx} 
\usepackage{hyperref} 
\usepackage{amsmath}  
\usepackage{amssymb}  
\usepackage{geometry}
\usepackage{float}
\usepackage{placeins}
\usepackage{subcaption}
\usepackage{enumitem}
\usepackage{comment}
\usepackage{amsmath}

\DeclareMathOperator*{\argmin}{arg\,min}

\providecommand{\keywords}[1]
{
  \small	
  \textbf{\textit{Keywords---}} #1
}

\title{Generative modeling of conditional probability distributions on the level-sets of collective variables}

\author{
Fatima-Zahrae Akhyar\thanks{ENPC, Institut Polytechnique de Paris, Marne-la-Vallée, France; Email: fatima-zahrae.akhyar@eleves.enpc.fr}
\and Wei Zhang\thanks{Zuse Institute Berlin, Takustrasse 7, Berlin 14195, Germany; Email: zhang@zib.de}
\and Gabriel Stoltz\thanks{CERMICS, ENPC, Institut Polytechnique de Paris, Marne-la-Vallée, France; MATHERIALS team-project, Inria Paris, France; Email: gabriel.stoltz@enpc.fr}
\and Christof Schütte\thanks{Institute of Mathematics, Freie Universität Berlin, Arnimalle 6, Berlin 14195, Germany; Zuse Institute Berlin, Takustrasse 7, Berlin 14195, Germany; Email: schuette@zib.de}
}

\date{}

\begin{document}
\maketitle

\begin{abstract}
   Given a probability distribution $\mu$ in $\mathbb{R}^d$ represented by data, we study in this paper the generative modeling of the corresponding conditional probability distributions on the level-sets of a collective variable $\xi: \mathbb{R}^d\rightarrow\mathbb{R}^k$, where $1\le k < d$. We propose a general and efficient learning approach that can learn generative models on different level-sets of $\xi$ simultaneously. To improve the learning quality on level-sets in low-probability regions, we also propose a data enrichment strategy by utilizing data from enhanced sampling techniques. We demonstrate the effectiveness of our proposed learning approach through concrete numerical examples. The proposed approach is potentially useful for the generative modeling of molecular systems in biophysics.   
\end{abstract}


\section{Introduction}
\label{sec-intro}

High-dimensional stochastic processes are ubiquitous in broad research areas such as biophysics, materials science, and climate modeling. Understanding the complex dynamical behavior of such processes often requires projecting the system down to a (much) lower-dimensional space defined by a few features, i.e.\ collective variables (CVs) or reaction coordinates, of the system. In molecular dynamics, for instance, various enhanced sampling techniques have been developed to increase the sampling efficiency and to construct the free energy surface in the space of CVs~\cite{enhanced-sampling-for-md-review}. 
A number of analytical and computational approaches have also been proposed for constructing simpler, lower-dimensional surrogate models associated to a chosen set of CVs of the original high-dimensional system. A notable approach is the effective dynamics constructed using conditional expectations, where the coefficients (i.e.\ drift and diffusion) in the surrogate model are computed by taking averages with respect to the so-called conditional probability distribution on the corresponding level-set of the CVs~\cite{effective_dynamics}.  
This approach has attracted considerable research interest in the mathematical community~\cite{ effective_dyn_2017, sz-entropy-2017, LEGOLL2017-pathwise,LelievreZhang18, 
LegollLelievreSharma18, 
DLPSS18,effdyn-transfer}, contributing to a deep understanding on the quality of the effective dynamics in approximating the original high-dimensional processes.

A common task in both of the aforementioned research directions, i.e.\ free energy calculation and the computation of coefficients in the effective dynamics, is sampling on the level-sets of CVs~\cite{projection_diffusion,Hartmann2008,Tony-constrained-langevin2012}. 
Metropolis-Hastings Monte Carlo methods have been developed for unbiased sampling on submanifolds defined by the level-sets of CVs~\cite{goodman-submanifold,hmc-submanifold-tony,
multiple-projection-2020}. Given a (Boltzmann--Gibbs) distribution in the full space, sampling methods for the corresponding conditional distribution on the level-sets of CVs have been developed by constructing ergodic diffusion processes on the level-sets~\cite{zhang2017,NonRevSam-2020}. More broadly, methods for sampling under constraints have also been studied in computational statistics~\cite{
Tavare1997,
Marin2012}.

In recent years, generative modeling approaches, in particular diffusion models~\cite{pmlr-v37-sohl-dickstein15,
NEURIPS2020_ddpm,
song2019generative,song2021scorebased} and flow-matching models~\cite{lipman2023flow,
LiuG023,
albergo2023building}, have achieved significant success in a wide range of real-world applications. 
Motivated by these achievements, there has been growing research interest in extending generative modeling approaches for Euclidean data to manifold data~\cite{mathieu2020riemannian,rozen2021moser,
BenHamu2022MatchingNF,de2022riemannian,
huang2022riemannian,
jo2024generativemodelingmanifoldsmixture,
chen2024flow,
rddpm,
liu2025improvingeuclideandiffusiongeneration}. Most of these existing approaches rely on substantial geometric information such as geodesics~\cite{de2022riemannian}, heat kernel, or eigenfunctions~\cite{chen2024flow}. As a result, their applicability is limited to manifolds where such information is explicitly known. Only the methods proposed in~\cite{rddpm,
liu2025improvingeuclideandiffusiongeneration} can be applied to manifolds that are implicitly defined as a level-set of a certain function. However, to the best of our knowledge, all these approaches are developed for generative modeling on a single and fixed manifold. 

In this paper, we propose an effective generative modeling approach for the conditional probability distributions on the level-sets of CVs. Our approach builds upon the flow-matching model~\cite{lipman2023flow,
LiuG023,
albergo2023building}, in which a neural ordinary differential equation (ODE) is learned to transport a prior (Gaussian) distribution to the target conditional distributions on the level-sets of a CV map $\xi$. Our method does not require explicit geometric information of the underlying manifolds and is broadly applicable to a wide range of manifolds that are implicitly defined by CVs. By varying the target CV value, the method can generate samples across different level-sets of $\xi$, rather than being restricted to a single level-set. To further improve the learning in regions of low probability density, we introduce a data enrichment strategy that leverages data obtained from enhanced sampling techniques such as adaptive biasing force (ABF) simulations. The approach is validated on a hierarchy of systems of increasing complexity, from synthetic low-dimensional examples to high-dimensional molecular datasets. The main contributions of this paper are summarized below:
\begin{enumerate}
    \item[(i)]   \textbf{A general and efficient generative modeling framework} capable of learning conditional distributions on families of manifolds implicitly defined by CVs;
    \item[(ii)]  \textbf{A data enrichment strategy} that utilizes enhanced sampling data to improve accuracy in low-density regions; and
    \item[(iii)] \textbf{Comprehensive numerical validations} that demonstrate the flexibility, robustness, and practical applicability of the method.
\end{enumerate}

The remainder of the paper is organized as follows. In Section~\ref{sec-notion}, we present the theoretical background and introduce the relevant notation. Section~\ref{sec-method} details the proposed generative modeling methodology. In Section~\ref{sec-examples}, we apply our approach to a series of representative datasets to illustrate its performance and generality. Finally, Section~\ref{sec-conclusion} concludes the paper and outlines possible directions for future research.

\section{Background and notation}
\label{sec-notion}
The aim of this section is to introduce the probability distributions of interest in this work, in particular the conditional distributions on the level-sets of a given smooth map.

Consider a distribution $\mu$ 
over $\mathbb{R}^d$ with a positive probability density $\rho$ with respect to the Lebesgue measure, namely $\mu(dx)=\rho(x)\,dx$. Assume that a $C^2$-smooth map $\xi: \mathbb{R}^d\rightarrow\mathbb{R}^k$ is given, where $1\le k< d$. 
We are particularly interested in applications in statistical physics, where $\mu$ arises as the invariant distribution of certain ergodic diffusion process $x_t$ in $\mathbb{R}^d$ and $\xi$ corresponds to the CV of the system. In this case, we often refer to the Boltzmann--Gibbs distribution, whose density with respect to the Lebesgue measure is 
\begin{equation}
\rho(x) = \frac{1}{Z} \mathrm{e}^{-\beta V(x)},\quad x\in \mathbb{R}^d\,,
\label{boltzman-mu}
\end{equation}
where $V:\mathbb{R}^d\rightarrow \mathbb{R}$ is a smooth potential function that grows sufficiently fast to infinity as $|x|\rightarrow +\infty$, $\beta$~is a positive constant whose inverse is proportional to the system's temperature, and $Z=\int_{\mathbb{R}^d} \mathrm{e}^{-\beta V(x)}\,dx$ is the normalizing constant that we assume to be finite. Below we introduce a few quantities related to the measure $\mu$ and the map $\xi$. Further discussions can be found in \cite[Section 3.2.1]{lelievre2010free} and \cite{effective_dyn_2017}. 

Given $z\in \mathbb{R}^k$, the level-set of the map $\xi$ corresponding to $z$ is defined as $$\Sigma_z:=\{x\in \mathbb{R}^d\,|\,\xi(x)=z\}.$$ Assume that $\Sigma_z$ is non-empty and that 
 $\nabla\xi(x)\in \mathbb{R}^{d\times k}$, the Jacobian of $\xi$, has full rank $k$ at each $x\in \Sigma_z$. Under these assumptions, the regular value theorem asserts that $\Sigma_z$ is a $(d-k)$-dimensional submanifold of~$\mathbb{R}^d$; see \cite[Chapter 1, Theorem 3.2]{hirsch2012differential}.
 Let $\widetilde{\mu}$ denote the pushforward  measure of $\mu$ by the map $\xi$; that is, $\widetilde{\mu}$ is the probability distribution of the random variable $\xi(X)\in \mathbb{R}^k$ when $X\sim \mu$. Let $\mu_z$ denote the conditional probability distribution on $\Sigma_z$, i.e. the probability distribution of $X$ conditioned on the event that $\xi(X)=z$. The law of total expectation implies 
\begin{equation}
\mathbb{E}_{X\sim\mu}(f(X)) = \mathbb{E}_{z \sim \widetilde{\mu}}\Big[\mathbb{E}_{X \sim \mu} \Big(f(X) \,\Big|\, \xi(X) = z\Big)\Big] 
= \mathbb{E}_{z \sim \widetilde{\mu}}\Big[\mathbb{E}_{X \sim \mu_z} \big(f(X)\big)\Big]\,,
\label{identity-total-expectation}
\end{equation}
 for test functions $f: \mathbb{R}^d\rightarrow\mathbb{R}$. Moreover, the co-area formula implies that (see \cite[Theorem 3.11]{evansmeasure} and \cite[Lemma~3.2]{lelievre2010free})
\begin{equation}
\mu_z(dx) = \frac{1}{Q(z)}\rho(x) \det \big(\nabla\xi(x)^\top\nabla\xi(x)\big)^{-\frac{1}{2}} \sigma_{\Sigma_z}(dx)\,, \quad x \in \Sigma_z\,,
\label{mu-z}
\end{equation}
and 
\begin{equation}
\widetilde{\mu}(dz) = Q(z)\,dz\,,\quad z \in \mathbb{R}^k\,,
\label{mu-eff-1}
\end{equation}
where  $\sigma_{\Sigma_z}$ denotes the surface measure  of $\Sigma_z$ and
\begin{equation}
 Q(z) = \int_{\Sigma_z} \rho(x) \det \big(\nabla\xi(x)^\top\nabla\xi(x)\big)^{-\frac{1}{2}} \sigma_{\Sigma_z}(dx)\,,\quad z \in \mathbb{R}^k\,,
 \label{factor-Q}
\end{equation}
is a normalizing factor in \eqref{mu-z}. 
For processes whose invariant distribution is the Boltzmann--Gibbs distribution $\mu(dx)=\frac{1}{Z} \mathrm{e}^{-\beta V(x)} dx$, the free energy associated to the map $\xi$ is defined as 
  \begin{equation}
F(z) = -\beta^{-1} \ln Q(z)\,,\quad z \in \mathbb{R}^k\,,
\label{fe}
\end{equation}
and, from \eqref{mu-eff-1}, we have
 \begin{equation}
\widetilde{\mu}(dz) = \mathrm{e}^{-\beta F(z)}\,dz\,,\quad z \in \mathbb{R}^k\,.
\label{mu-eff-2}
\end{equation}

\section{Learning approach}
\label{sec-method}
In this section, we present our methodology for learning generative models on the level-sets of a smooth map $\xi$. We first describe the standard approach based on flow-matching for learning the conditional distributions $\mu_z$ corresponding to different values $z\in \mathbb{R}^k$. Then, we propose an enhanced learning strategy that leverages biased trajectory data to improve the quality of the learning in regions of low probability under the pushforward distribution $\widetilde{\mu}$. Finally, we explain how such biased trajectory data can be generated in practice using adaptive biasing techniques.

\subsection{Standard learning approach}
\label{subsec-method-standard}
Our goal is to sample the conditional probability distributions $\mu_z$ for different values $z\in \mathbb{R}^k$. For this purpose, we adopt the flow-matching generative modeling framework~\cite{LiuG023,lipman2023flow,albergo2023building}, which employs the ordinary differential equation (ODE) in $\mathbb{R}^d$ 
\begin{equation}
\frac{dX_t}{dt} = v(X_t,t)\,,  \quad X_0\sim \nu_{\mathrm{prior}}\,,\quad t\in [0,1]
\label{fm-ode}
\end{equation}
to transform a prior distribution $\nu_{\mathrm{prior}}$ to a target distribution $\nu$ that is represented by data. For this purpose, the vector field $v$ in \eqref{fm-ode} is learned by minimizing the standard flow-matching objective~\cite{LiuG023} 
\begin{equation}
    \mathcal{L}_{\mathrm{FM}, \nu}(v) =\mathbb{E}_{t\sim U([0,1])} \mathbb{E}_{X_0\sim \nu_{\mathrm{prior}},X_1\sim \nu} \left\| v \big((1-t)X_0+tX_1, t\big)- (X_1 - X_0)\right\|^2 \,,
    \label{fm-loss}
\end{equation}
where $U([0,1])$ denotes the uniform distribution on $[0,1]$.
It is known that the minimizer of \eqref{fm-loss} leads to an ODE \eqref{fm-ode} that transforms the prior distribution $\nu_{\mathrm{prior}}$ to the target $\nu$. 

Adapting this framework to our setting, we propose to learn the ODE  
\begin{equation}
\frac{dX_t}{dt} = v^{(z)}(X_t, t)\,,  \quad X_0\sim \nu_{\mathrm{prior}}\,,\quad t\in [0,1],
\label{fm-ode-z}
\end{equation}
parametrized by a variable $z\in \mathbb{R}^k$, 
such that, for a fixed $z\in \mathbb{R}^k$, the system \eqref{fm-ode-z} transforms the prior distribution from $t=0$ to the conditional distribution $\mu_z$ at $t=1$. To achieve this goal, we consider the following objective for learning the vector field in \eqref{fm-ode-z}:
\begin{equation}
 \mathcal{L}(v)= \mathbb{E}_{t\sim U([0,1])}\mathbb{E}_{X_0\sim \nu_{\mathrm{prior}},X_1\sim \mu} \left\| v^{(\xi(X_1))} \big((1-t)X_0+tX_1, t\big)- (X_1 - X_0)\right\|^2 \,,
 \label{fm-loss-xi-unbiased}
\end{equation}
where $\mu$ is the probability distribution introduced in the previous section. To justify \eqref{fm-loss-xi-unbiased}, notice that we can derive 
\begin{equation}
\begin{aligned}
 \mathcal{L}(v)&= \mathbb{E}_{X_1\sim \mu}\mathbb{E}_{t\sim U([0,1]), X_0\sim \nu_{\mathrm{prior}}}\left\| v^{(\xi(X_1))} \big((1-t)X_0+tX_1, t\big)- (X_1 - X_0)\right\|^2 \\
 &= \mathbb{E}_{z\sim \widetilde{\mu}} \left[\mathbb{E}_{t\sim U([0,1])}\mathbb{E}_{X_0\sim \nu_{\mathrm{prior}}, X_1\sim \mu_z} \left\| v^{(z)} \big((1-t)X_0+tX_1, t\big)- (X_1 - X_0)\right\|^2\right] \\
 &= \mathbb{E}_{z\sim \widetilde{\mu}}\left( \mathcal{L}_{\mathrm{FM}, \mu_z}(v^{(z)})\right)\,,
 \end{aligned}
 \label{fm-loss-xi-unbiased-rewrite}
\end{equation}
where the second equality follows from the law of total expectation in~\eqref{identity-total-expectation}.
 This implies that minimizing the objective \eqref{fm-loss-xi-unbiased} is equivalent to minimizing the standard flow-matching objective \eqref{fm-loss} simultaneously for different target distributions $\mu_z$. Therefore, it allows us to learn the ODE \eqref{fm-ode-z} for generative modeling on level-sets $\Sigma_z$ corresponding to different values $z$ at the same time. 

 In practice, \eqref{fm-loss-xi-unbiased} is replaced by the empirical objective  
\begin{equation}
 \mathcal{L}^{(N)}(v)= \frac{1}{N} \sum_{n=1}^N \left\| v^{(\xi(X^{(n)}))} \big((1-t_n)X_0^{(n)}+t_nX^{(n)}, t\big)- (X^{(n)} - X^{(n)}_0)\right\|^2 \,,
 \label{fm-loss-xi-unbiased-empirical}
\end{equation}
where $t_1, t_2, \dots, t_N$ are drawn independently from $U([0,1])$,  $X_0^{(1)}, X_0^{(1)}, \dots, X_0^{(N)}$ are drawn independently from the prior $\nu_{\mathrm{prior}}$, and $X^{(1)}, X^{(2)}, \dots, X^{(N)}$ are the data points. 

\subsection{Enhanced learning approach with biased trajectory data}
\label{subsec-method-enhance}

We consider the case where the target distribution $\mu$ is the Boltzmann--Gibbs distribution with a density in \eqref{boltzman-mu}. Let us introduce the overdamped Langevin dynamics 
\begin{equation}
dx_t = - \nabla V(x_t)\, dt + \sqrt{2\beta^{-1}}\, dw_t\,,\quad t\ge 0\,,
\label{overdamped}
\end{equation}
where $V:\mathbb{R}^d\rightarrow \mathbb{R}$ is a smooth potential function that increases to infinity sufficiently fast as $|x|\rightarrow +\infty$ and $w_t$ is a standard Brownian motion in $\mathbb{R}^d$. 
It is known that $\mu$ is invariant under the process $x_t$ in \eqref{overdamped} and that $x_t$ is ergodic with respect to $\mu$.
Thanks to the ergodicity of the process, the empirical objective \eqref{fm-loss-xi-unbiased-empirical} can be evaluated using trajectory data $X^{(1)}, X^{(2)}, \dots, X^{(N)}$ obtained by sampling \eqref{overdamped} with a numerical scheme at time $t=h, 2h, \dots, Nh$, where $h$ denotes the step-size and $N$ is the total number of sampling steps. 

We are particularly interested in the case where the potential $V$ has multiple local minima and $\beta$ is relatively large (i.e.\ the temperature is low). In this case, the probability density of the invariant distribution $\mu$ is highly non-uniform in space, since the probability that $x_t$ is within the vicinity of one of the local minima (metastable regions) is much larger than the probability that $x_t$ is found elsewhere. Likewise, for a proper CV map $\xi$, the pushforward distribution $\widetilde{\mu}$ defined in \eqref{mu-eff-1} is highly non-uniform in $\mathbb{R}^k$ as well. 
The generative model that is learned by minimizing the objective \eqref{fm-loss-xi-unbiased} (or its empirical counterpart \eqref{fm-loss-xi-unbiased-empirical}) using training data sampled from \eqref{overdamped} 
may be inferior in the low-density regions under $\widetilde{\mu}$, because of the relatively small
contribution of the loss in such regions to the aggregated objective \eqref{fm-loss-xi-unbiased} (see \eqref{fm-loss-xi-unbiased-rewrite}). 

In order to overcome the aforementioned issue, we utilize the observation (as can be seen from the derivation \eqref{fm-loss-xi-unbiased-rewrite}) that different probability distributions of $X_1$ can be chosen in \eqref{fm-loss-xi-unbiased}, as long as the corresponding conditional probability distributions coincide with the conditional probability distributions $\mu_z$. In particular, let us consider the reweighted (RW) probability distribution 
\begin{equation}
\mu_{\mathrm{RW}}(dx) = \bar{Z}^{-1} Z^{-1}\mathrm{e}^{-\beta \big(V(x)- A(\xi(x))\big)}\,dx\,,\quad x\in \mathbb{R}^d\,,
\label{mu-rw}
\end{equation}
where $A:\mathbb{R}^k\rightarrow \mathbb{R}$ and $\bar{Z}=Z^{-1}\int_{\mathbb{R}^d} \mathrm{e}^{-\beta \big(V(x)- A(\xi(x))\big)}\,dx$ is a normalizing factor.
Then, for each $z\in \mathbb{R}^k$, it is straightforward to verify that the conditional probability distribution of $\mu_{\mathrm{RW}}$ on the level-set $\Sigma_z$ coincides with $\mu_z$ given in \eqref{mu-z}. Moreover, using \eqref{factor-Q} and \eqref{fe}, we can obtain that the pushforward of $\mu_{\mathrm{RW}}$ by the map $\xi$ satisfies
\begin{equation}
\widetilde{\mu}_{\mathrm{RW}}(dz) = \bar{Z}^{-1} \mathrm{e}^{-\beta \big(F(z)-A(z)\big)} dz\,,\quad z\in \mathbb{R}^k\,,
\label{mu-1-eff}
\end{equation}
where $F$ is the free energy \eqref{fe} associated to the map $\xi$.
 With the choice of the probability distribution $\mu_{\mathrm{RW}}$ in \eqref{mu-rw}, the objective \eqref{fm-loss-xi-unbiased} changes to
\begin{equation}
 \mathcal{L}_{\mathrm{RW}}(v)= \mathbb{E}_{t\sim U([0,1])}\mathbb{E}_{X_0\sim \nu_{\mathrm{prior}},X_1\sim \mu_{\mathrm{RW}}} \left\| v^{(\xi(X_1))} \big((1-t)X_0+tX_1, t\big)- (X_1 - X_0)\right\|^2 \,,
 \label{fm-loss-xi-rw}
\end{equation}
and similar to \eqref{fm-loss-xi-unbiased-rewrite} we can derive that  
\begin{equation}
\begin{aligned}
 \mathcal{L}_{\mathrm{RW}}(v)=& \mathbb{E}_{z\sim \widetilde{\mu}_{\mathrm{RW}}}\Big( \mathcal{L}_{\mathrm{FM}, \mu_z}(v^{(z)})\Big)\,.
 \end{aligned}
 \label{fm-loss-xi-rw-rewrite}
\end{equation}
From \eqref{mu-1-eff}, we can see that the density of $\widetilde{\mu}_{\mathrm{RW}}$ is more uniform than the density of $\widetilde{\mu}$ in \eqref{mu-eff-2} when $A$ is close to the free energy $F$. Therefore, \eqref{fm-loss-xi-rw-rewrite} implies that in this case
the contribution of the loss for values $z$ in low-density regions under $\widetilde{\mu}$ is enhanced in the objective \eqref{fm-loss-xi-rw}. 

To generate training data, we notice that $\mu_{\mathrm{RW}}$ in \eqref{mu-rw} is the invariant distribution of the process
\begin{equation}
dx_t = - \nabla (V-A\circ\xi)(x_t)\,dt + \sqrt{2\beta^{-1}}\,dw_t\,,\quad t\ge 0\,.
\label{overdamped-rw}
\end{equation}
To ensure that $A$ approximates the free energy $F$, we adopt the adaptive biasing force (ABF) method~\cite{abf-darve2008,abf-namd,Fiorin2013,abf-everything,EABF}. Concretely, we generate training data by simulating the biased process 
\begin{equation}
dx_t = - \nabla V(x_t)\,dt +  \nabla\xi(x_t) f_t(\xi(x_t))\,dt + \sqrt{2\beta^{-1}}\,dw_t\,,\quad t\ge 0\,,
\label{overdamped-rw-1}
\end{equation}
where $f_t(z)$ is an estimation of the gradient of the free energy $F$ (i.e.\ the mean force), based on the expression~\cite[Lemma 3.9]{lelievre2010free}:
\begin{equation}
\frac{\partial F}{\partial z_i}(z) = \mathbb{E}_{\mu_z}\left[\sum_{l=1}^{d}\sum_{j=1}^{k}
(\nabla \xi^\top\nabla\xi)^{-1}_{ij} \frac{\partial \xi_j}{\partial x_l}\frac{\partial V}{\partial x_l} -  \beta^{-1}\sum_{l=1}^{d}\frac{\partial}{\partial x_l}\bigg(\sum_{j=1}^{k}(\nabla \xi^\top\nabla\xi)^{-1}_{ij} \frac{\partial \xi_j}{\partial x_l}\bigg) \right]\,,~ i=1,2,\dots, k\,,
\label{mean-force}
\end{equation}
where $(\nabla \xi^\top \nabla \xi)^{-1}_{ij}$ denotes the $ij$-component of the inverse of the matrix $\nabla \xi^\top \nabla \xi\in \mathbb{R}^{k\times k}$. In practice, the mean value \eqref{mean-force} is estimated on a grid in $\mathbb{R}^k$. In each simulation step, the estimation is updated for the cell where the current state $\xi(x_t)$ belongs to. After the cell has been visited for a certain (pre-defined) number of times, the current value of the estimation in that cell is used as the biasing term $f_t(\xi(x_t))$ in \eqref{overdamped-rw-1}. As $t\rightarrow +\infty$, the empirical estimations are expected to converge to the true means in \eqref{mean-force} and, accordingly, the process \eqref{overdamped-rw-1} is expected to converge to its equilibrium counterpart \eqref{overdamped-rw} with $A=F$. 

\subsection{Projection step}
\label{subsec-projection}
Once the model is learned using the approach described in the previous section, it can be used to generate new data points on different level-sets of $\xi$ by integrating the ODE~\eqref{fm-ode-z} with different values $z\in \mathbb{R}^k$.
However, due to numerical errors, the generated data may not lie exactly on the target level-set. Here, we propose a post-processing step that can be used optionally (depending on the specific applications) to project the generated data onto the target level-set. Also refer to~\cite{liu2025improvingeuclideandiffusiongeneration} for discussions on projection onto manifolds in the context of generative modeling. In the following, we discuss two different projection methods. The first method is the projection to the closest point on the level-set, which amounts to solving 
\begin{equation}
    \bar{x} := \argmin_{y\in \Sigma_z} \|x - y\|^2
    \label{projection-closest}
\end{equation}
for each generated point $x$. However, unless the level-set $\Sigma_z$ can be parametrized by certain coordinates, solving \eqref{projection-closest} is  computationally infeasible because it involves constraints. The second method is to project the generated state $x$ along the ODE flow~\cite{zhang2017}
\begin{equation}
    \frac{dY_t}{dt} = - \nabla G(Y_t)\,,\quad Y_0=x\,,
\label{projection-ode-flow}
\end{equation}
where $G(y):=\frac{1}{2}\|\xi(y) - z\|^2$ for $y\in \mathbb{R}^d$.
  Under certain regularity assumptions on~$\xi$, the quantity $G(Y_t)$,  as $t\rightarrow +\infty$, decreases to zero exponentially fast thanks to the gradient structure of \eqref{projection-ode-flow},
  implying that the state evolved by the flow of~\eqref{projection-ode-flow} approaches the level-set $\Sigma_z$. 
In practice, to perform the projection, it suffices to integrate the ODE \eqref{projection-ode-flow} 
up to a time when certain criteria is met, e.g.\ $\|\xi(Y_t)-z\|< \epsilon_{\mathrm{tol}}$, where 
$\epsilon_{\mathrm{tol}}>0$ is a small parameter specifying the tolerance.

It is worth pointing out that, while this projection step is helpful to meet the condition that the states lie on the target level-set, it does not necessarily improve the quality of the generated data, as measured by the distance between the empirical distribution of the data and the target conditional probability distribution on the level-set. For systems in physics, it may also produce projected states that are not physically meaningful. To avoid such issues, it is necessary that the generative model is well trained, so that the generated data (before projection) is close to the target level-sets. Besides, incorporating geometrical or physical information in the projection step (e.g.\ by requiring that the distance between the projected state and the starting state to be small or the energy to be low) could be helpful. Finally, instead of a projection step, one could also relax the data points generated on the target submanifold with some constrained Langevin dynamics in order to improve their physical relevance~\cite{Tony-constrained-langevin2012,hmc-submanifold-tony}.

\section{Experiments}
\label{sec-examples}
In this section, we demonstrate the capability of our method by applying it to different representative datasets. First, we consider a synthetic dataset in $\mathbb{R}^2$ to illustrate the method. Next, we evaluate the performance of our method on the trajectory data of the process under the Müller--Brown potential, which is a widely used benchmark system exhibiting metastability. Finally, we apply our method to the alanine dipeptide molecular system. The neural networks and the training were implemented using \texttt{PyTorch}~\cite{paszke2019pytorch}.

\subsection{2D Synthetic Dataset}

\begin{figure}[ht!]
    \centering
    \begin{subfigure}[b]{0.45\textwidth}  
        \centering
	\includegraphics[width=0.82\textwidth]{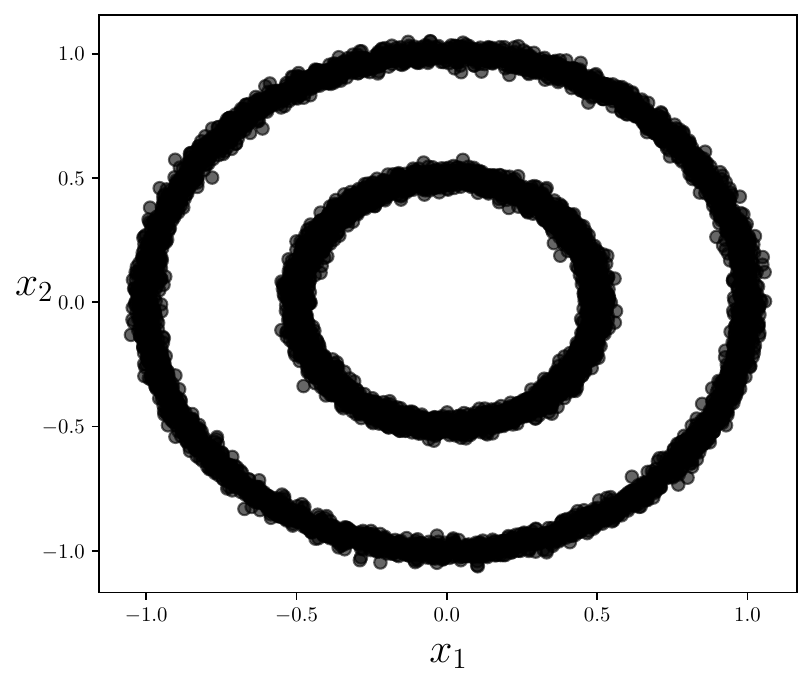}  
        \caption{\label{fig:2d-data-a}}
    \end{subfigure}
    \hspace{1cm}
    \begin{subfigure}[b]{0.45\textwidth}  
        \centering
	\includegraphics[width=0.82\textwidth]{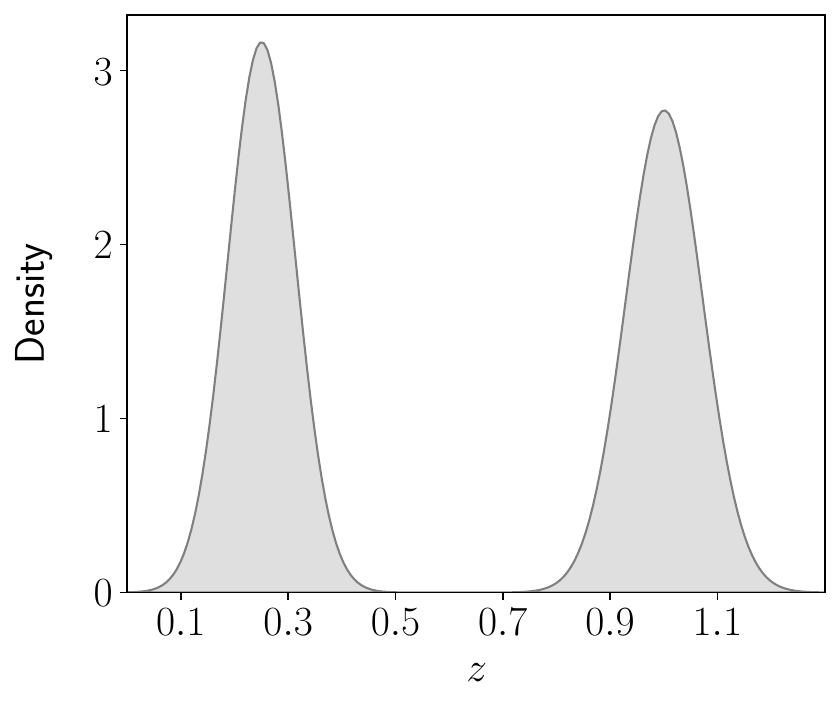} 
        \caption{\label{fig:2d-data-b}}
    \end{subfigure}
    \caption{(a) Scatter plot of the 2D dataset sampled using the \texttt{make\_circles} function from \texttt{scikit-learn}~\cite{scikit-learn}. 
    (b) Density estimate of the corresponding CV map $\xi(x) = x_1^2 + x_2^2$.}
    \label{fig:2d_data_cv}
\end{figure}

\begin{figure}[ht!]
    \centering
    \begin{subfigure}[b]{0.32\textwidth}
        \centering
        \includegraphics[width=0.95\textwidth]{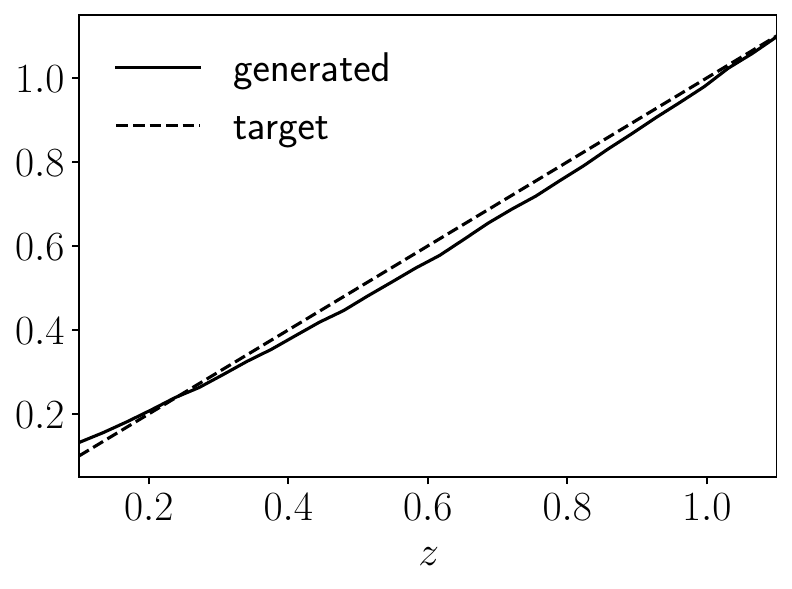}
        \caption{\label{fig:2d-result-a}}
    \end{subfigure}
    \begin{subfigure}[b]{0.32\textwidth}
        \centering
        \includegraphics[width=0.95\textwidth]{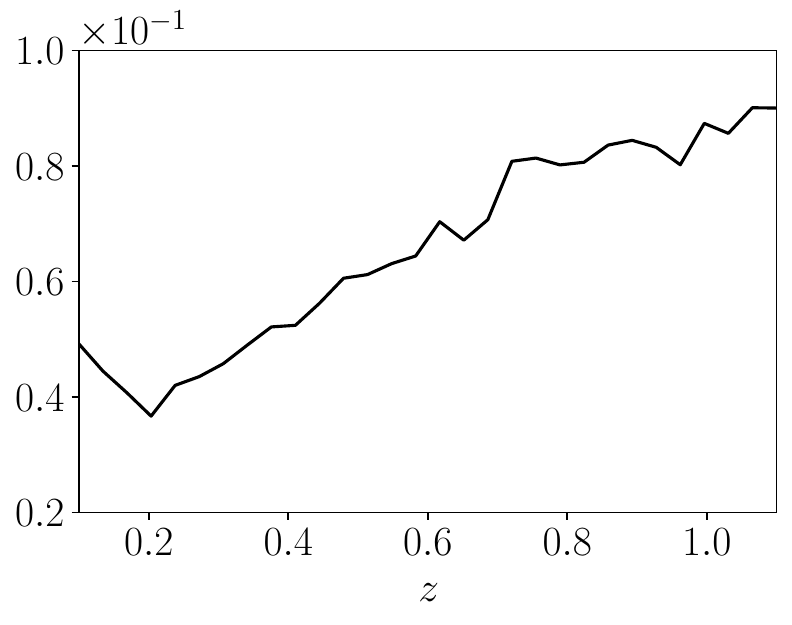}
        \caption{\label{fig:2d-result-b}}
    \end{subfigure}    
    \begin{subfigure}[b]{0.32\textwidth}
        \centering
        \includegraphics[width=0.95\textwidth]{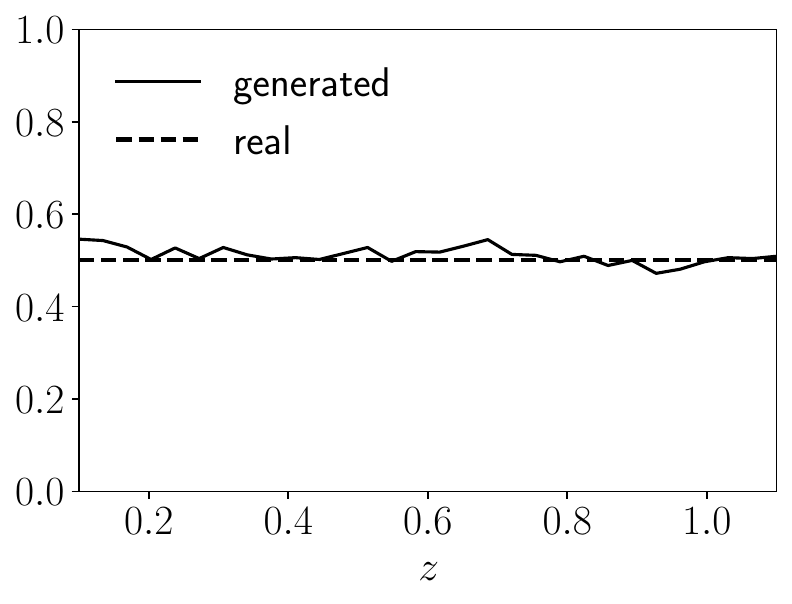}
        \caption{\label{fig:2d-result-c}}
    \end{subfigure}

    \caption{Results for the 2D dataset. (a) Mean value of $\xi$ on the generated samples compared to the intended CV value. 
    (b) Mean deviation of $\xi$ on generated samples from the target CV value. 
  (c) Proportion of samples with positive $x_1$ values in the original and in the generated samples for different target CV values.}     
    \label{fig:2d_all}
\end{figure}

\begin{figure}[ht!]
    \centering
    \includegraphics[width=0.9\textwidth]{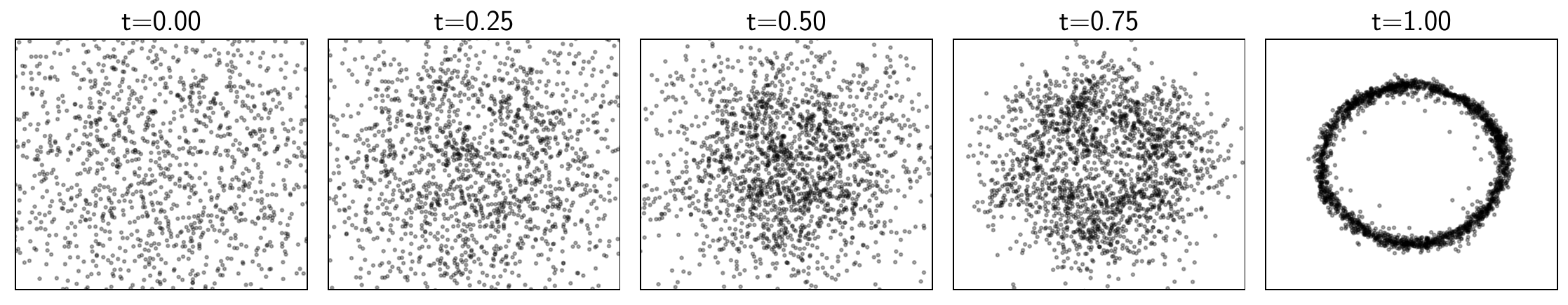}
    \caption{Evolution of the two-dimensional samples under the ODE flow for a fixed target CV value $z=0.6$. The panels show the particle positions at different integration times $t$, illustrating how the initial Gaussian cloud progressively morphs into the target distribution.}
    \label{fig:2d_evolution}
\end{figure}

We study the 2D dataset (Figure~\ref{fig:2d-data-a}) that is sampled using the \texttt{make\_circles} function from the package \texttt{scikit-learn}~\cite{scikit-learn}. The CV map is defined as $\xi(x) = x_1^2 + x_2^2$ for $x=(x_1,x_2)\in \mathbb{R}^2$, whose empirical density is shown in Figure~\ref{fig:2d-data-b}. 

The prior $\nu_{\mathrm{prior}}$ is chosen as the standard Gaussian distribution in $\mathbb{R}^2$.
The vector field $v$ in \eqref{fm-ode-z} is parameterized as a fully connected feedforward neural network, comprising a total of four layers: an input layer, two hidden layers with $128$ neurons each and \texttt{Tanh} activations, followed by an output layer of size $2$. The input consists of the spatial coordinate $x$ in $\mathbb{R}^2$, a time scalar $t$, and the target CV value $z$ (of dimension 1). 
To learn the vector field $v$, the empirical objective $\mathcal{L}^{(N)}(v)$ in \eqref{fm-loss-xi-unbiased-empirical} was minimized using the Adam optimizer for up to 1000 epochs with early stopping (patience of 200 epochs), a learning rate of $10^{-3}$, a weight decay of $10^{-4}$, and a batch size of $1000$. For other parameters, their default values in \texttt{PyTorch} were used. 

Next, we evaluate the performance of the learned model for generating states conditioned on the value $z$ of the map $\xi$. To this end, we choose the values of $z$ uniformly within an extended range around the values visited in the dataset (with a 5\% margin), and for each chosen value of $z$, $1000$ new states are generated by integrating the ODE \eqref{fm-ode-z} using the \texttt{torchdiffeq} library \cite{chen2018neural} (4th-order Runge-Kutta scheme) up to $t=1$ with $1000$ time steps, starting from the standard Gaussian prior at $t=0$.
Based on the generated states, we estimate the mean value of $\xi$, the mean deviation of $\xi$ from the target CV value $z$, defined as 
\begin{equation}
\mathrm{Deviation}(z) = \sqrt{\frac{1}{N}\sum_{i=1}^{N} \Big(\xi(X^{(i)}) - z\Big)^2},
\label{def-deviation-z-paragraph}
\end{equation}
where $(X^{(i)})_{1\le i \le N}$ are the generated samples,
as well as the proportion of states with $x_1\ge 0$, which are shown in Figures~\ref{fig:2d-result-a},~\ref{fig:2d-result-b}, and~\ref{fig:2d-result-c}, respectively.
The results in Figure~\ref{fig:2d_all} demonstrate that the trained model performs consistently well across all evaluation metrics. Namely, the generated CV values closely match the intended targets, the deviation remains low across the entire range of values, and the model accurately learns the conditional distribution.

Finally, Figure~\ref{fig:2d_evolution} presents the temporal evolution of the generated distribution for $z=0.6$, obtained by integrating the ODE from an initial cloud of $2000$ points sampled from the standard Gaussian prior. 
Successive snapshots are taken at increasing integration times $t \in \{0, 0.25, 0.50, 0.75, 1.0\}$, showing how the initially isotropic Gaussian distribution gradually transforms into the target ring-shaped configuration.

\subsection{Müller--Brown Dataset}

\begin{figure}[ht!]
    \centering
    \includegraphics[width=0.5\textwidth]{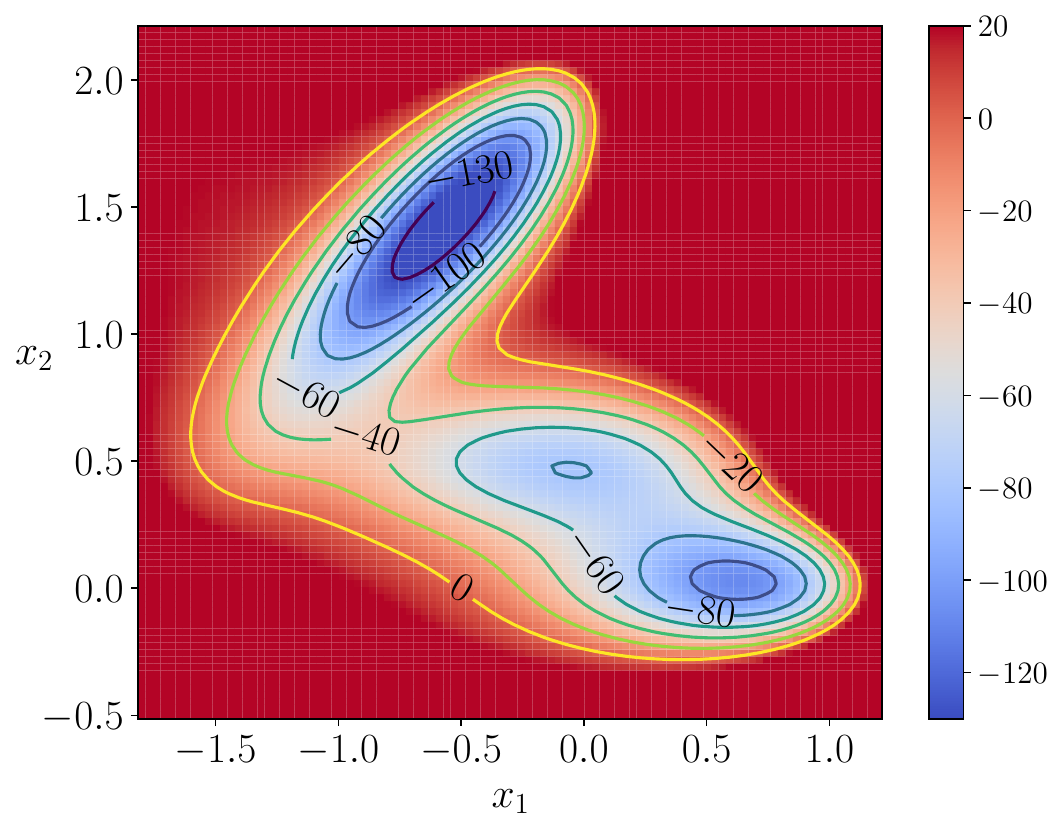}
    \caption{Heat map and contour representation of the Müller--Brown potential landscape. 
    The two major low-potential regions and the shallow low-potential region are shown in dark blue and light blue, respectively, while the contour lines highlight the specific potential energy levels.}
    \label{fig:Muller}
\end{figure}

We apply our approach to the two-dimensional Boltzmann--Gibbs distribution~\eqref{boltzman-mu} with the Müller--Brown potential~\cite{muller1979location}. As shown in Figure~\ref{fig:Muller}, the potential has two deep wells (i.e.\ low-potential regions), separated by a shallow well. At low temperatures, the corresponding process \eqref{overdamped} exhibits metastability with a curved transition pathway, making it an ideal model system for studying the complex behavior of metastable dynamics.  

\paragraph{Model Description and Training.}
The dataset was prepared by sampling the trajectory of the overdamped Langevin dynamics \eqref{overdamped} up to time $t=600$
(corresponding to $3\times10^6$ steps), with the integration time step $\Delta t = 2 \times 10^{-4}$ and the inverse temperature $\beta = 0.1$, starting from the state $(-0.6, 1.2)$ in the deepest well. A dataset consisting of $3\times10^4$ states was obtained by recording the states every 100 steps.

\begin{figure}[t!]
    \centering
    \begin{subfigure}[b]{0.45\textwidth}
        \centering
        \includegraphics[width=0.9\textwidth]{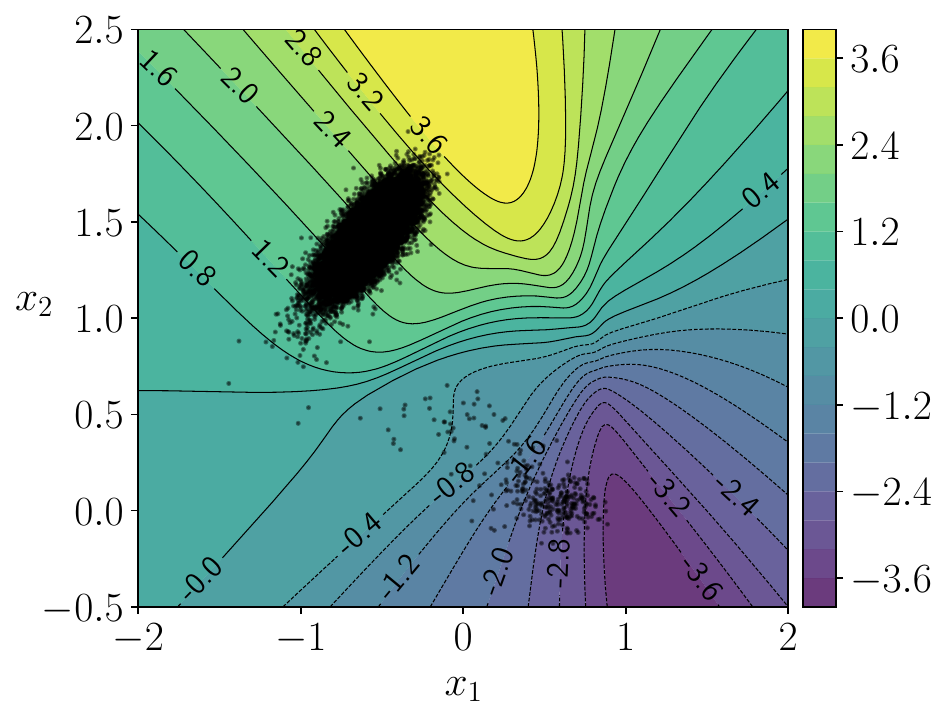}
        \caption{}
        \label{fig:MB-original-a}
    \end{subfigure}
    \begin{subfigure}[b]{0.46\textwidth}
        \centering
        \includegraphics[width=0.9\textwidth]{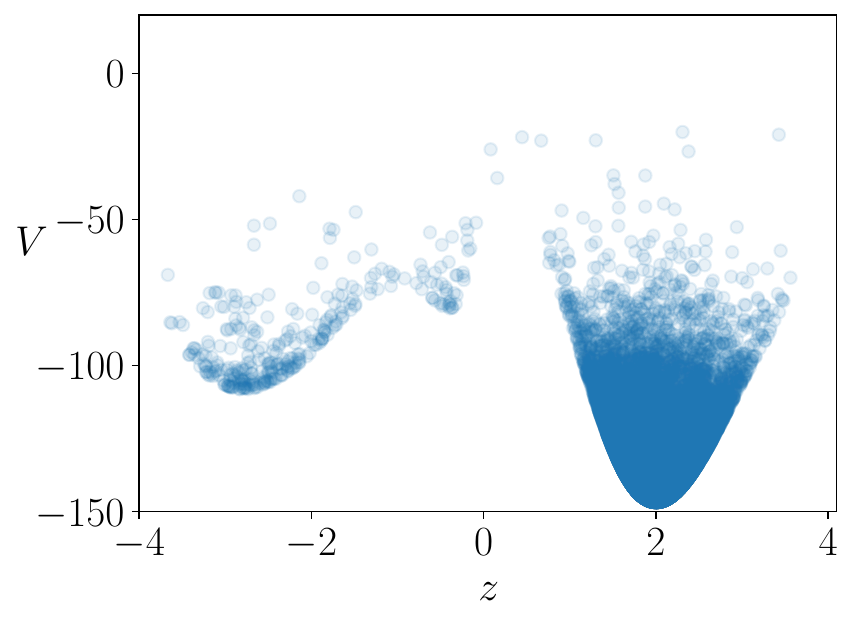}
        \caption{}
        \label{fig:MB-original-b}
    \end{subfigure}
    \caption{
    (a) Trajectory samples from unbiased sampling overlaid on level-sets of the learned CV map $\xi$. The color bar indicates the corresponding CV values across the configuration space. (b) Scatter plot of the potential energy versus the CV value.
    }
    \label{fig:MB_original_data}
\end{figure}

In order to define a CV map,  we first trained an autoencoder with a one-dimensional bottleneck, modeled as a fully connected feedforward neural network, by minimizing the mean squared reconstruction error on the generated trajectory data. The encoder network consists of an input layer of size 2, three hidden layers of size 32 with \texttt{Tanh} activation, and an output layer of size 1. The decoder network consists of an input layer of size 1, a single hidden layer of size 32 with \texttt{Tanh} activation, and an output layer of size 2. Training was performed for $200$ epochs using the Adam optimizer, with a learning rate of $10^{-3}$ and a batch size of~$128$. The CV map $\xi: \mathbb{R}^2\rightarrow \mathbb{R}$ is defined by the trained encoder. Figure~\ref{fig:MB-original-a} shows the level-sets of $\xi$ with the sampled trajectory data superimposed, whereas Figure~\ref{fig:MB-original-b} shows the scatter plot of the potential energy \(V(x)\) for each state $x$ in the trajectory against the encoded CV value $\xi(x)$.

\begin{figure}[ht!]
    \centering
    \begin{subfigure}[b]{0.45\textwidth}
        \centering
        \includegraphics[width=0.9\textwidth]{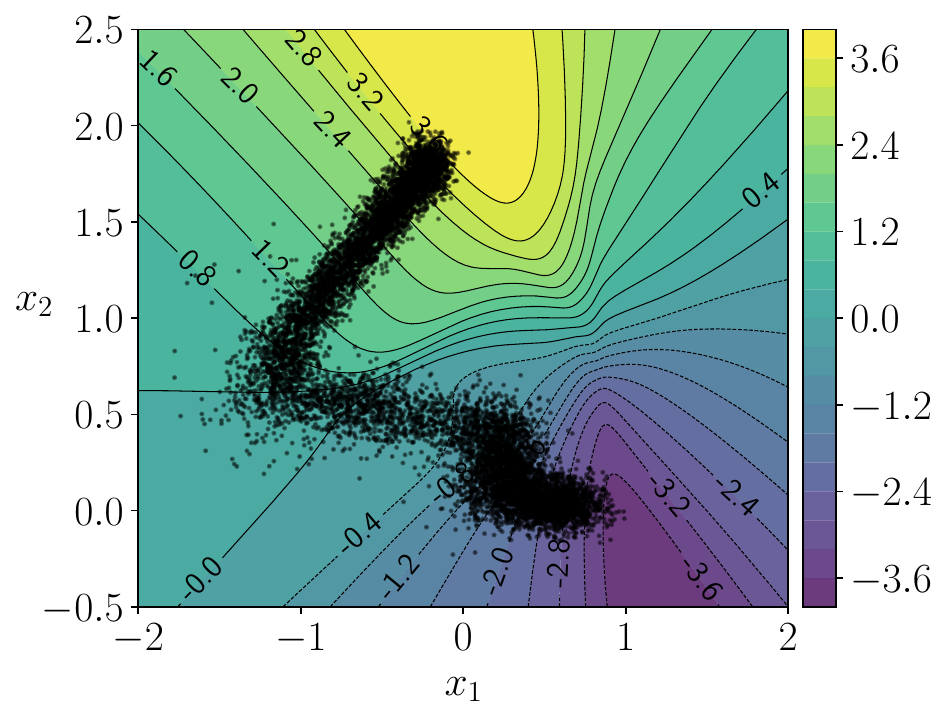}
        \caption{}
        \label{fig:MB-biased-a}
    \end{subfigure}
    \begin{subfigure}[b]{0.46\textwidth}
        \centering
        \includegraphics[width=0.9\textwidth]{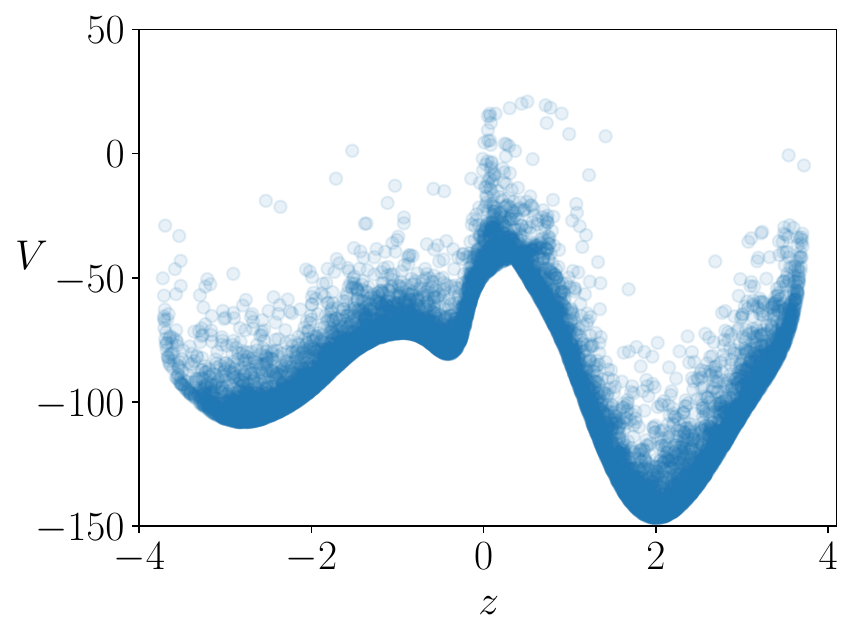}
        \caption{}
        \label{fig:MB-biased-b}
    \end{subfigure}
    \caption{(a) Trajectory samples obtained from the ABF simulation overlaid on level-sets of the learned CV map $\xi$. The color bar indicates the corresponding CV values across the configuration space. (b) Scatter plot of the potential energy versus the CV value for states in the ABF-biased trajectory.}
    \label{fig:MB_biased_data}
\end{figure}

To test the enhanced learning technique discussed in Section~\ref{subsec-method-enhance},  we also employed the ABF dynamics \eqref{overdamped-rw-1} with the learned CV map $\xi$, where the biasing force $f_t$ was estimated according to \eqref{mean-force} on a uniform grid with grid spacing $0.1$ that covers the range $[-2.6, 3.5]$. To ensure accurate estimation, the biasing force in a grid cell was applied to the system after the cell has been visited 100 times (equivalently, 100 samples have been collected to estimate \eqref{mean-force}). The biased simulation was run for $1.5 \times 10^5$ steps and, after the initial $5\times 10^4$ simulation steps (equilibration), states were recorded every $10$ steps, resulting in a trajectory dataset of size $10^4$.  
Similar to Figures~\ref{fig:MB-original-a}--\ref{fig:MB-original-b}, Figures~\ref{fig:MB-biased-a} and \ref{fig:MB-biased-b} show the scatter plot of the ABF-biased trajectory data superimposed on the level-sets of $\xi$ and the scatter plot of potential values $V(x)$ against the encoded CV value $\xi(x)$, respectively.

We trained the ODE model \eqref{fm-ode-z} using the generated trajectory data and the learned CV map $\xi$. The vector field $v$ is modeled by a fully connected feedforward neural network, which consists of an input layer of size 4 (i.e.\ the total dimension of the state $x\in \mathbb{R}^2$, time $t \in [0,1]$ and the target CV value $z\in \mathbb{R}$), three hidden layers of size 128 with \texttt{Tanh} activation, and an output layer of size $2$. Training was performed for 3000 epochs by minimizing the objective \eqref{fm-loss-xi-unbiased-empirical} using the Adam optimizer with a learning rate of $10^{-3}$ and a batch size of $512$.

\begin{figure}[t!]
    \centering
    \begin{subfigure}[b]{0.45\textwidth}
        \centering
        \includegraphics[width=0.9\textwidth]{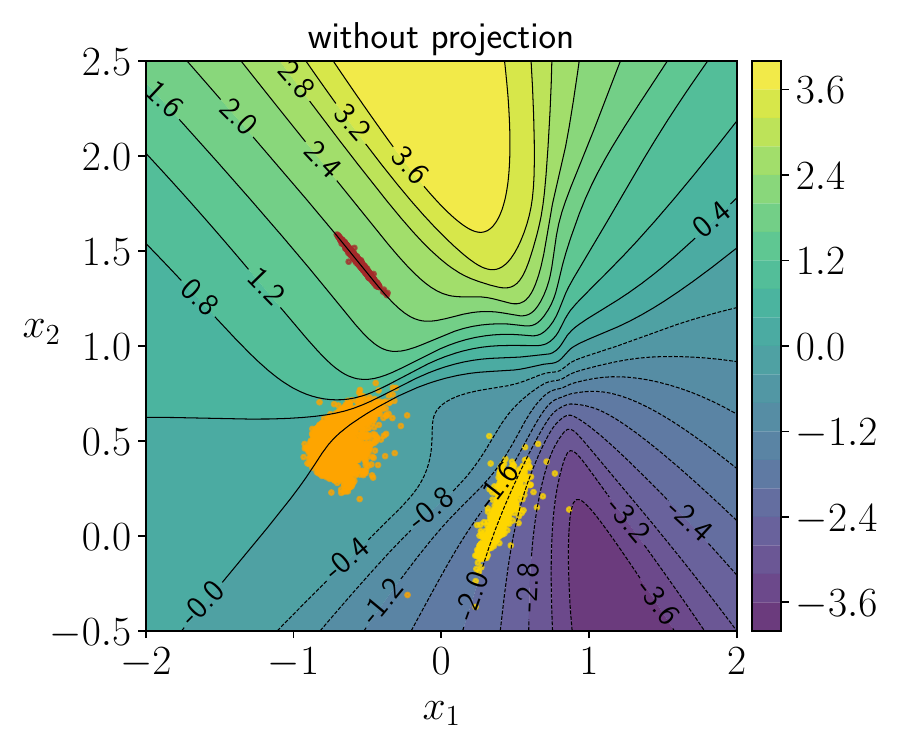}
        \caption{}
        \label{fig:MB-generated-a-unbiased}
    \end{subfigure}
    \begin{subfigure}[b]{0.45\textwidth}
        \centering
        \includegraphics[width=0.9\textwidth]{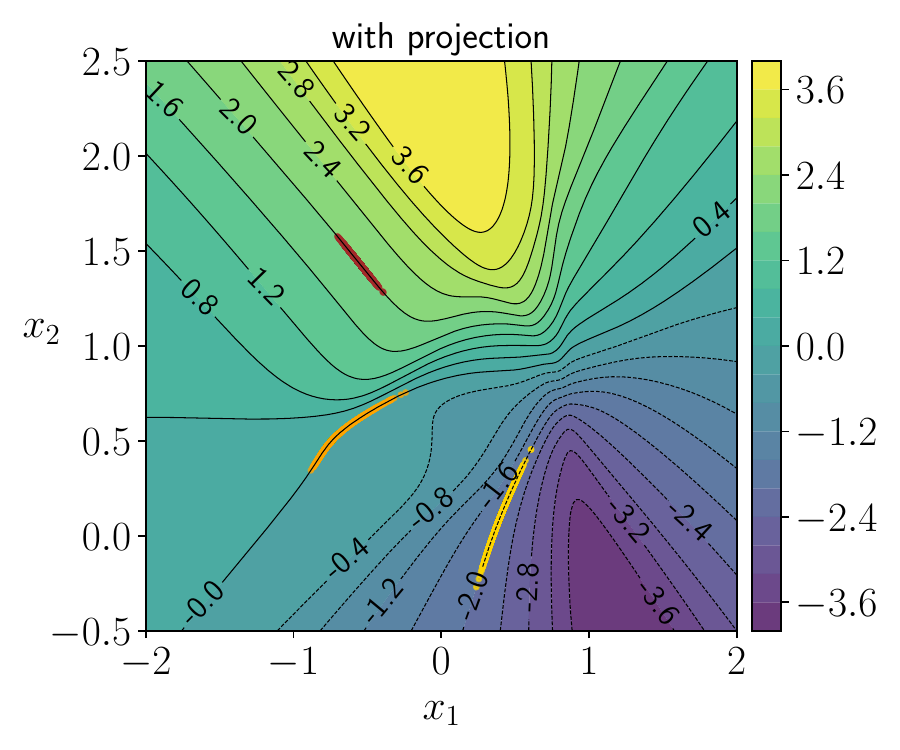}
        \caption{}
        \label{fig:MB-generated-b-unbiased}
    \end{subfigure}
    \caption{Generated samples for different CV values $z\in\{-2.0,0.0,2.0\}$ using the model trained on the unbiased trajectory data. (a) Without projection. (b) With projection.}
    \label{fig:MB_generated_proj_unbiased}
\end{figure}

\begin{figure}[t!]
    \centering
    \begin{subfigure}[b]{0.45\textwidth}
        \centering
        \includegraphics[width=0.9\textwidth]{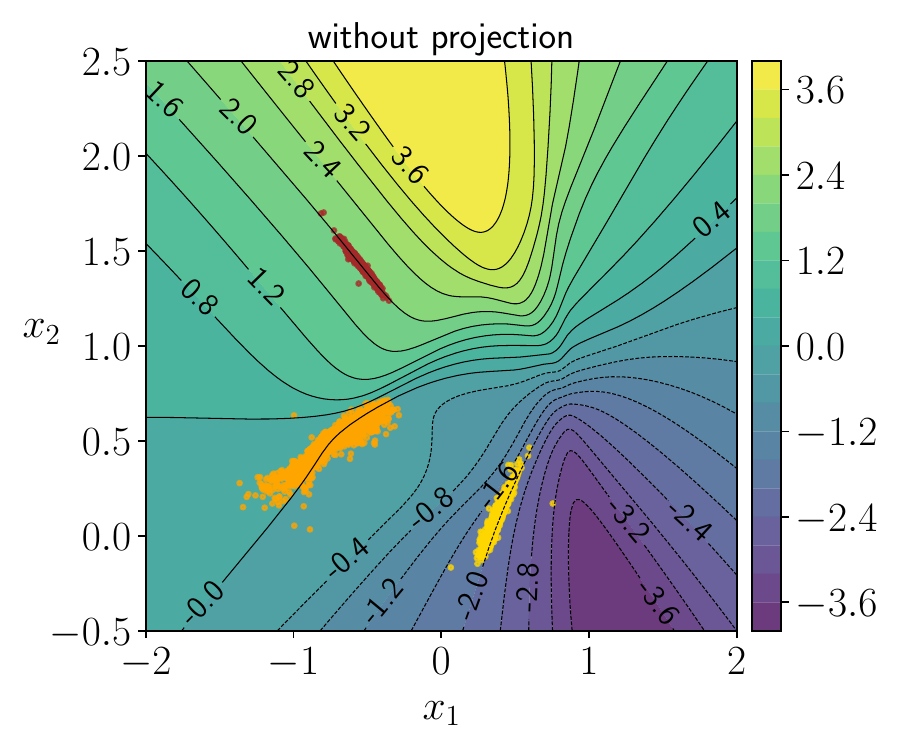}
        \caption{}
        \label{fig:MB-generated-a-biased}
    \end{subfigure}
    \begin{subfigure}[b]{0.45\textwidth}
        \centering
        \includegraphics[width=0.9\textwidth]{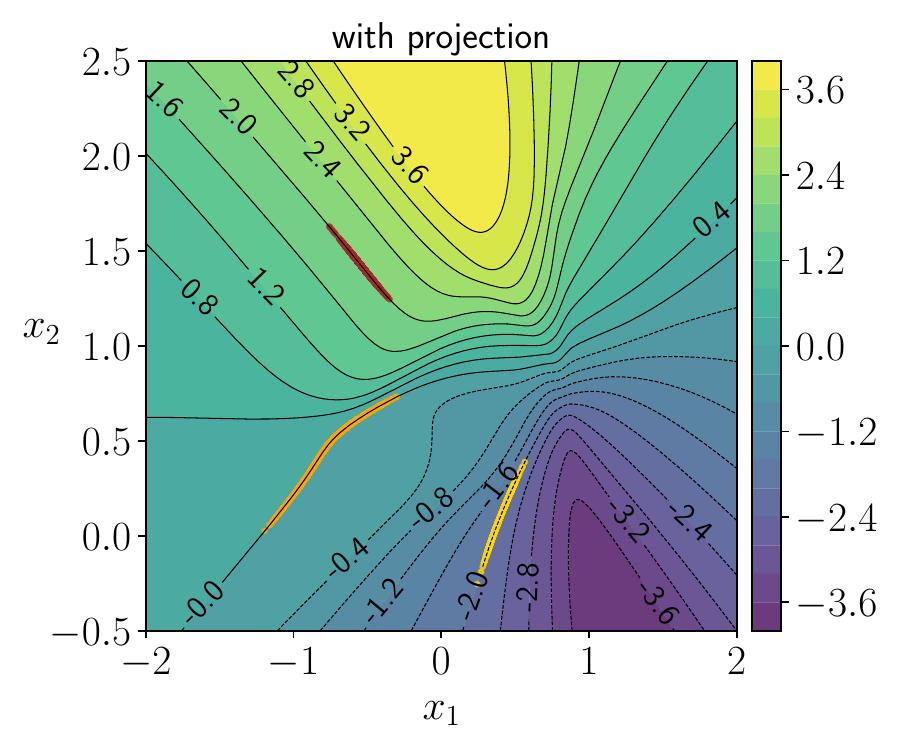}
        \caption{}
        \label{fig:MB-generated-b-biased}
    \end{subfigure}
    \caption{Generated samples for different CV values $z\in\{-2.0,0.0,2.0\}$ using the model trained on the ABF-biased trajectory data. (a) Without projection. (b) With projection.
    }
    \label{fig:MB_generated_proj_biased}
\end{figure}

\begin{figure}[ht!]
    \centering
    \includegraphics[width=0.35\textwidth]{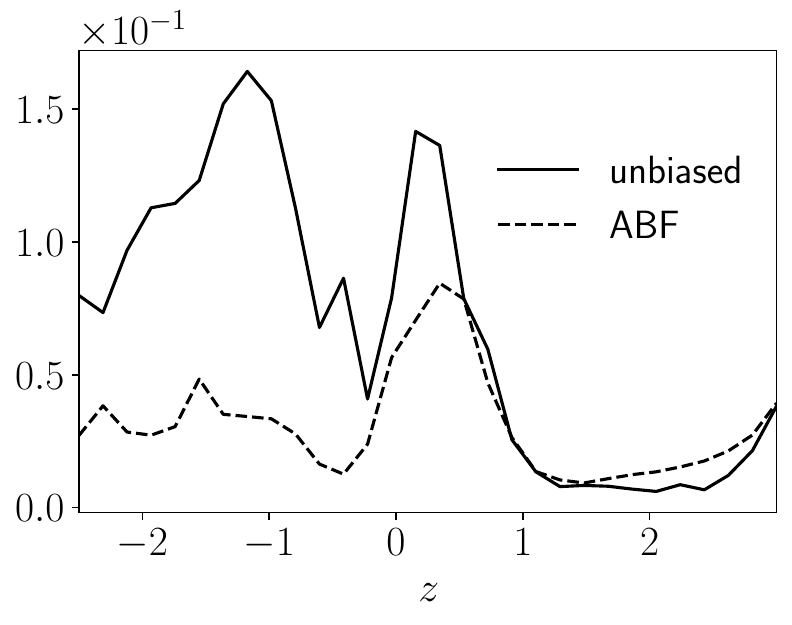}
    \caption{The deviation defined in~\eqref{def-deviation-z-paragraph} as a function of the target CV value for the model trained on the unbiased trajectory data and the model trained on the ABF-biased trajectory data.}
    \label{fig:proj-c}
\end{figure}

\begin{figure}[ht!]
    \centering
    \includegraphics[width=0.95\textwidth]{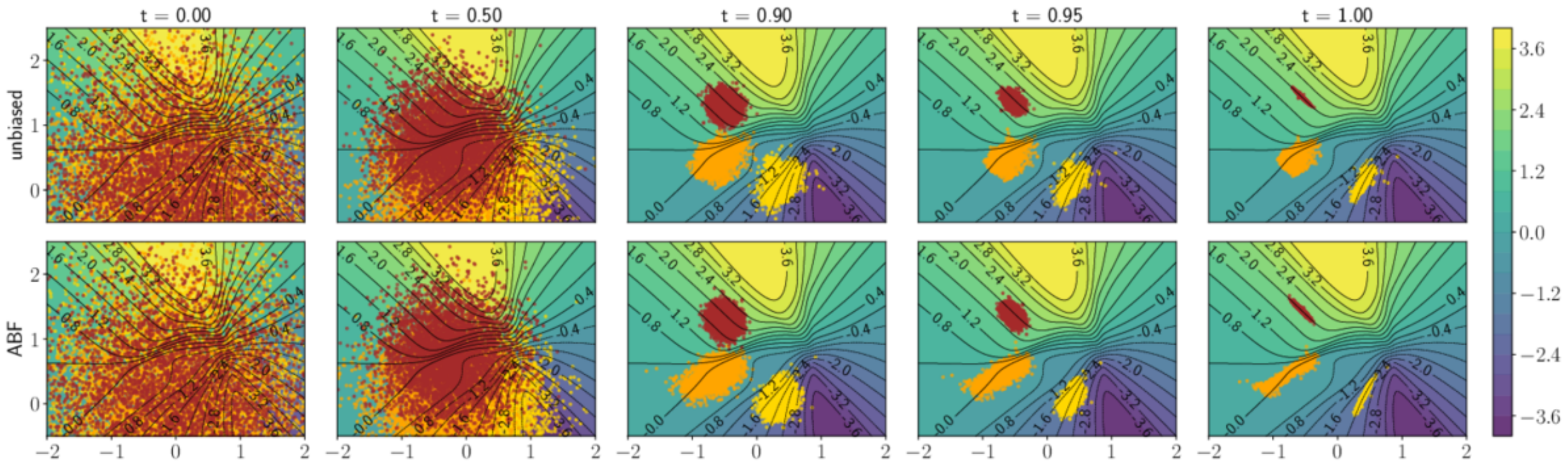} 
    \caption{Time evolution of the data points under the learned model \eqref{fm-ode-z} for three different target values of $z\in \{-2.0,0.0,2.0\}$. The top part of the figure corresponds to the model trained with the unbiased trajectory data, and the bottom part corresponds to the model trained with the ABF-biased trajectory data. Contours and filled colormaps show the level-sets and the profile of the CV map $\xi$, respectively. Points corresponding to each $z$ value are drawn in distinct colors to track their evolution over time.}
    \label{fig:flow_evolution}
\end{figure}

\paragraph{Results.}
We evaluated the trained generative models by generating new states with prescribed target CV values~$z\in \{-2.0,0.0, 2.0\}$. For each target CV value $z$, $10^3$ states sampled from the two-dimensional standard Gaussian distribution were propagated by numerically integrating the ODE
\eqref{fm-ode-z} using the
\texttt{torchdiffeq} library \cite{chen2018neural} (4th-order Runge-Kutta scheme) up to $t=1$ with $100$ uniform time steps. The resulting final states at time $t=1$ were interpreted as samples associated with the chosen CV value $z$. These samples are visualized in Figures~\ref{fig:MB-generated-a-unbiased} and \ref{fig:MB-generated-a-biased}. Comparing these two figures reveals that the model trained using the ABF-biased trajectory data produces spatially consistent samples distributed along the correct CV contours (Figure~\ref{fig:MB-generated-a-biased}), whereas the model trained using the unbiased trajectory data tends to generate inconsistent samples when the target level-sets are not well covered by the training data (i.e.\ level-sets corresponding to $z=-2.0$ and $z=0.0$ in Figure~\ref{fig:MB-generated-a-unbiased}). To quantify the evaluation, the CV values of the generated samples were compared to the target value $z$, and the root mean squared deviation defined in \eqref{def-deviation-z-paragraph} was computed for the target values $z$ on a uniform grid of the interval $[-2.9, 3.6]$. The deviation curves in Figure~\ref{fig:proj-c} summarize this evaluation: the model trained on the ABF-biased trajectory achieves a lower deviation across the CV range. Figure~\ref{fig:flow_evolution} shows snapshots of the temporal evolution of the point distributions under the learned ODE model~\eqref{fm-ode-z}, taken at specific time points $t = 0.0, 0.5, 0.9, 0.95, 1.0$, illustrating the transformation from the initial standard Gaussian distribution to the final distribution conditioned on the latent variable $z$.

%%%%%%%%%%%%%% biased data

We also tested the method introduced in Section~\ref{subsec-projection} for projecting the generated samples. For each target CV value $z$, the samples obtained from the generative models 
(as shown in Figures~\ref{fig:MB-generated-a-unbiased} and~\ref{fig:MB-generated-a-biased}) were propagated by integrating the ODE~\eqref{projection-ode-flow} numerically using the explicit Euler scheme, where we used step size $0.01$, the maximal number of integration steps $7000$ and tolerance $\epsilon_{\mathrm{tol}}=10^{-3}$. The resulting states are shown in 
Figures~\ref{fig:MB-generated-b-unbiased} and \ref{fig:MB-generated-b-biased}, respectively, from which we can conclude that the generated samples were successfully projected onto the corresponding level-sets. 

\begin{figure}[ht!]
    \centering
    \begin{subfigure}[b]{0.47\textwidth}
        \centering
        \includegraphics[width=0.9\textwidth]{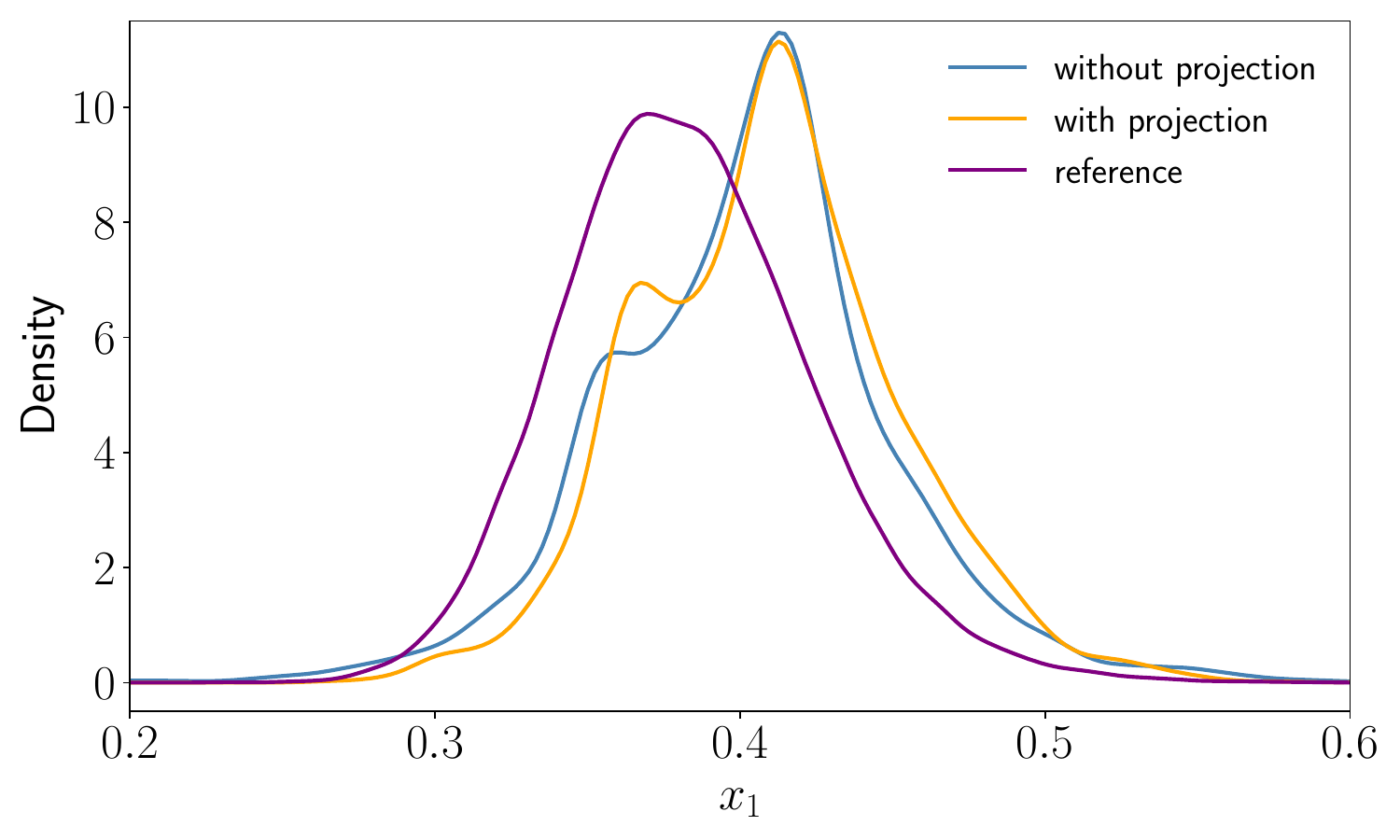}
        \caption{}
        \label{fig:proj-a}
    \end{subfigure}
    \begin{subfigure}[b]{0.47\textwidth}
        \centering
        \includegraphics[width=0.9\textwidth]{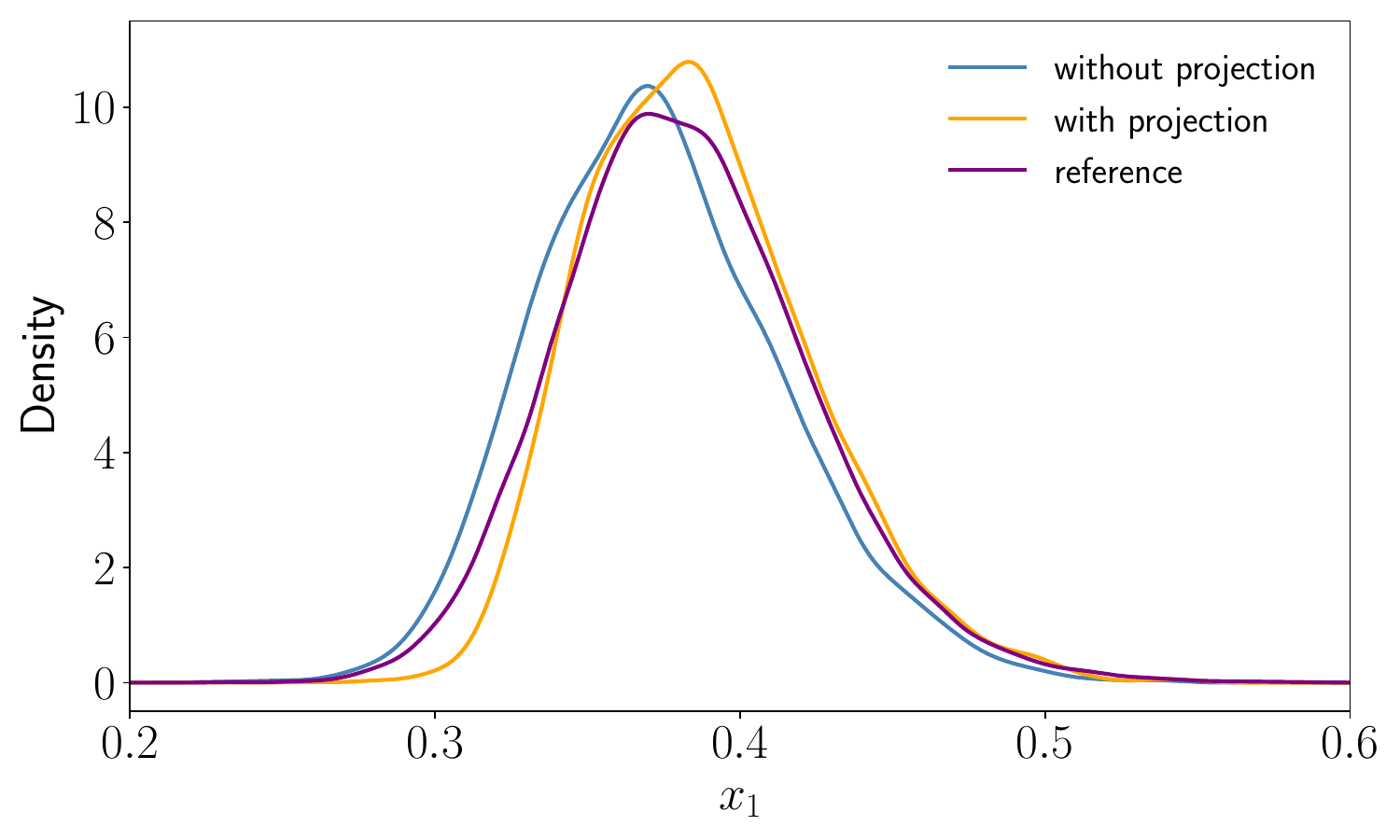}
        \caption{}
        \label{fig:proj-b}
    \end{subfigure}
    \caption{
    Density estimations of the first component $x_1$ for the samples generated using the generative model for the target CV value $z=-2.0$ (without projection), their projections onto the level-sets (with projection), and the samples obtained by constrained sampling on the level-set (reference). (a) Results for the generative model trained using the unbiased trajectory data. (b) Results for the generative model trained using the ABF-biased trajectory data. }
    \label{fig:before_after_projection}
\end{figure}

Finally, we assessed the closeness between the empirical distribution of the generated samples and the conditional distribution on the corresponding level-set. We chose $z=-2.0$ and sampled $5\times 10^4$ states on the corresponding level-set using the constrained sampling scheme in \cite{projection_diffusion} with a step size $2.0\times 10^{-4}$. In order to approximate the true conditional distribution, each of the sampled states $x$ was assigned the weight $|\nabla\xi(x)|^{-1}$, which is proportional to the likelihood ratio between the conditional distribution~\eqref{mu-z} and the invariant distribution of the constrained numerical scheme in the continuous-time limit (see \cite{projection_diffusion,zhang2017} for detailed discussions). 
With these data, we estimated the densities of the first component $x_1$, using the samples generated by the learned generative model, their projections on the level-set by the method in Section~\ref{subsec-projection}, and the weighted samples obtained by constrained sampling.
The results are shown in Figure~\ref{fig:proj-a} and Figure~\ref{fig:proj-b} for the model trained using the unbiased trajectory data and the model trained using the ABF-biased trajectory data, respectively. From these two figures, we can conclude that the projection maps the generated states onto the level-set without modifying their distribution significantly. Moreover, it can be observed that, compared to the model trained using the unbiased trajectory data, the ODE model trained using the ABF-biased trajectory generates data whose distribution is closer to the true conditional distribution. We also computed probability densities of other quantities such as the potential energy and the distance to the mean. The results are similar to those in Figure~\ref{fig:before_after_projection} and therefore are not shown here. 

\subsection{Alanine Dipeptide}

\paragraph{Model Description and Training.} We applied our
generative approach to alanine dipeptide, which is a small benchmark molecular
system consisting of 22 atoms. We studied two different generative modeling tasks. 
In the first task, our goal is to train the ODE model~\eqref{fm-ode-z} to generate  configurations of the system's 10 non-Hydrogen atoms conditioned on its two backbone dihedral angles $\phi$ and $\psi$. In the second task, our goal is to train the ODE model~\eqref{fm-ode-z} to generate configurations of the system's 10 non-Hydrogen atoms conditioned on the dihedral angle $\phi$ alone.

\begin{figure}[t!]
    \centering
    \begin{subfigure}[t]{0.32\linewidth}
    \centering
    \includegraphics[width=0.95\textwidth]{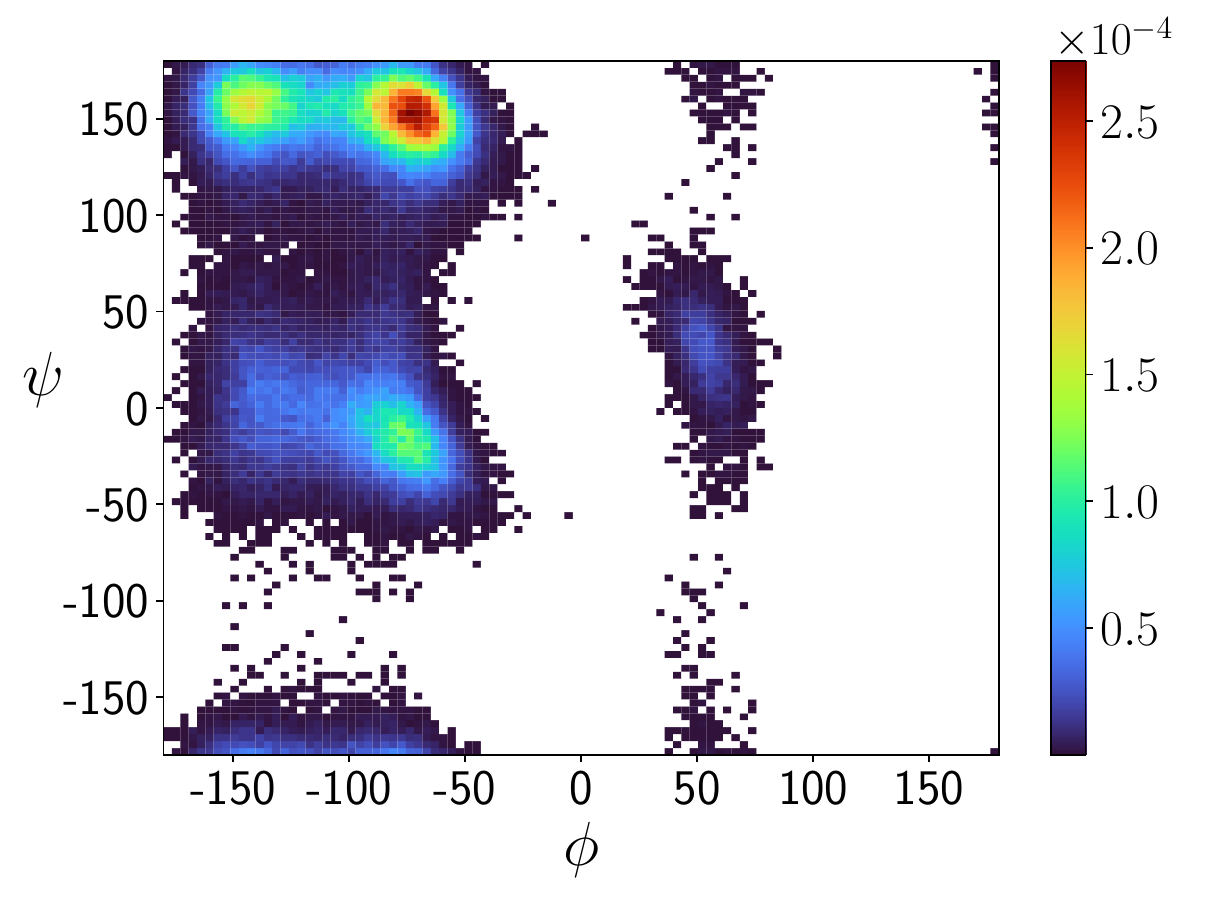}
    \caption{\label{fig:phi_psi_density_unbiased}}
    \end{subfigure}
    \centering
    \begin{subfigure}[t]{0.32\linewidth}
    \includegraphics[width=0.95\textwidth]{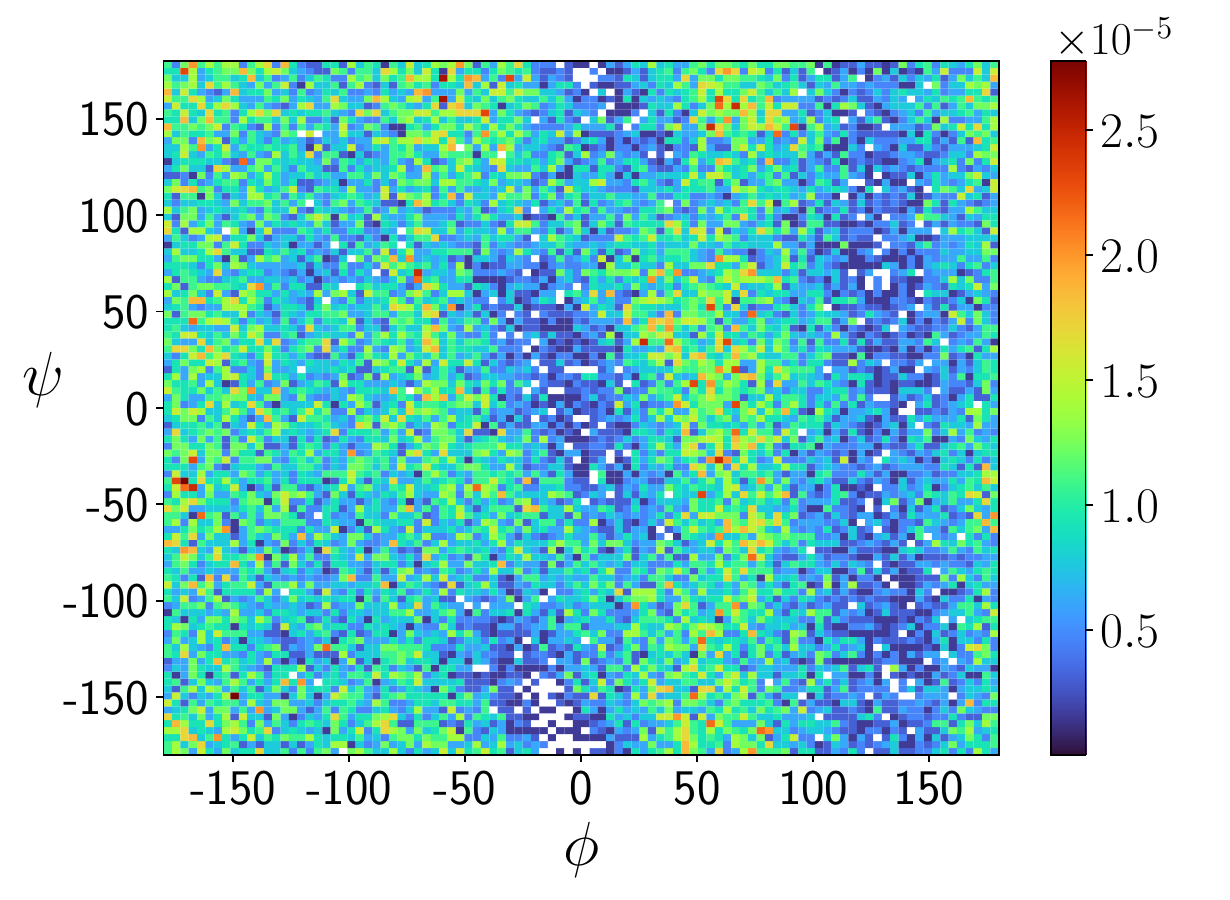}
    \caption{\label{fig:phi_psi_density_abf_cv2d}}
\end{subfigure}
   \begin{subfigure}[t]{0.32\linewidth}
   \centering
    \includegraphics[width=0.95\textwidth]{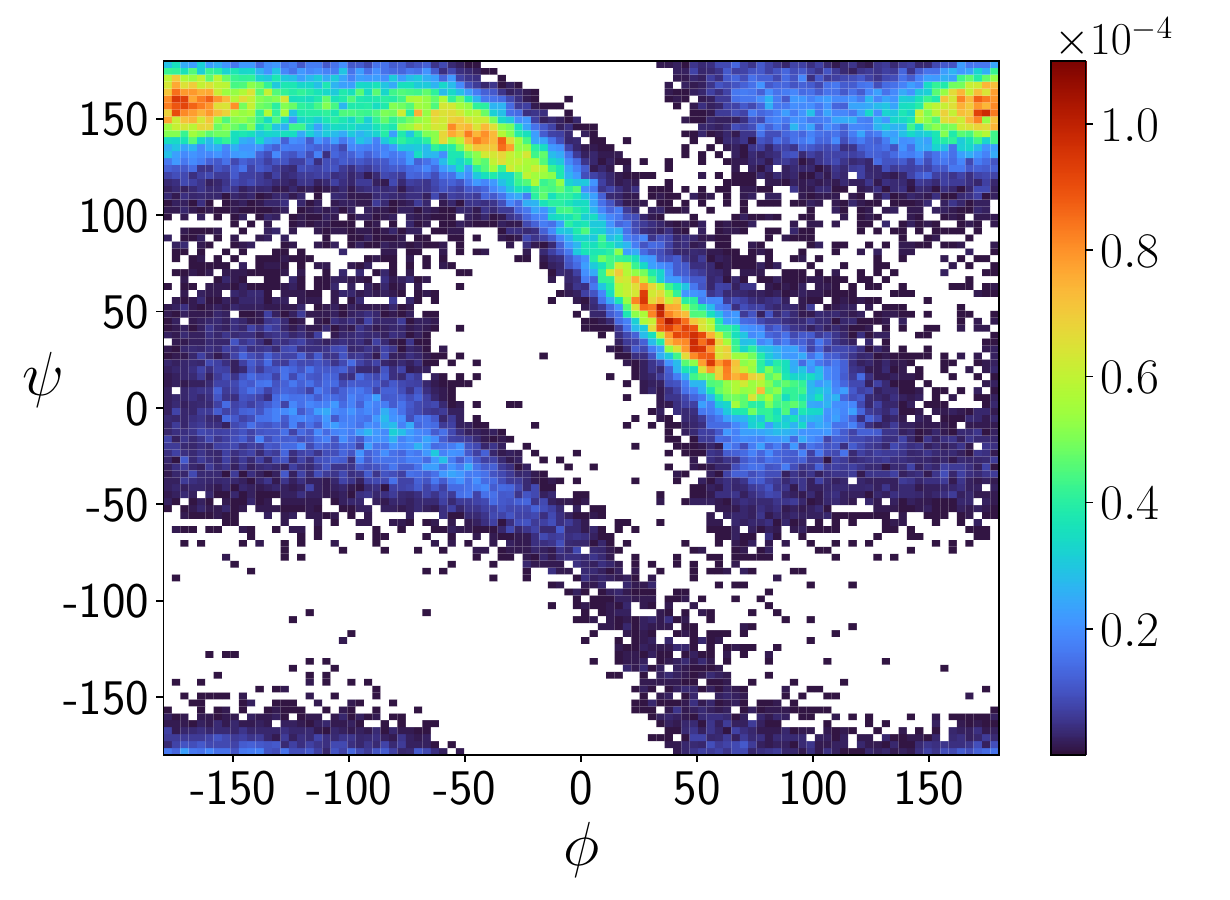}
    \caption{\label{fig:phi_psi_density_abf_cv1d}}
\end{subfigure}
\caption{Empirical densities of the dihedral angles of the trajectory data.
  (a) Trajectory from the unbiased simulation. (b) Trajectory from the ABF
  simulation, where the CV is defined as the dihedral angles $(\phi,\psi)$.
  (c) Trajectory from the ABF simulation, where the CV is defined as the dihedral angle $\phi$.}
\end{figure}

Trajectory data of the alanine dipeptide system in water was generated by performing molecular dynamics simulations for $t=1500\,\textrm{ns}$ using the GROMACS package~\cite{GROMACS}, where we adopted the same simulation parameters as in the previous work~\cite{lelievre2023analyzing} and the states were recorded every $10$\,ps, resulting in a trajectory dataset of $1.5\times 10^5$ states.
Cartesian coordinates of the 10 non-Hydrogen atoms were then extracted using MDAnalysis~\cite{gowers2019mdanalysis,michaud2011mdanalysis} and aligned to a reference configuration via the Kabsch algorithm~\cite{kabsch1978discussion}, thereby eliminating global translational and rotational motions.
The aligned coordinates were then standardized by subtracting the empirical mean and dividing by the empirical standard deviation before being used as training data.
Figure~\ref{fig:phi_psi_density_unbiased} shows the empirical density of the trajectory data projected onto the space of the two dihedral angles.

\begin{figure}[t!]
\centering
    \begin{subfigure}[t]{0.46\linewidth}
    \centering
      \includegraphics[width=0.98\textwidth]{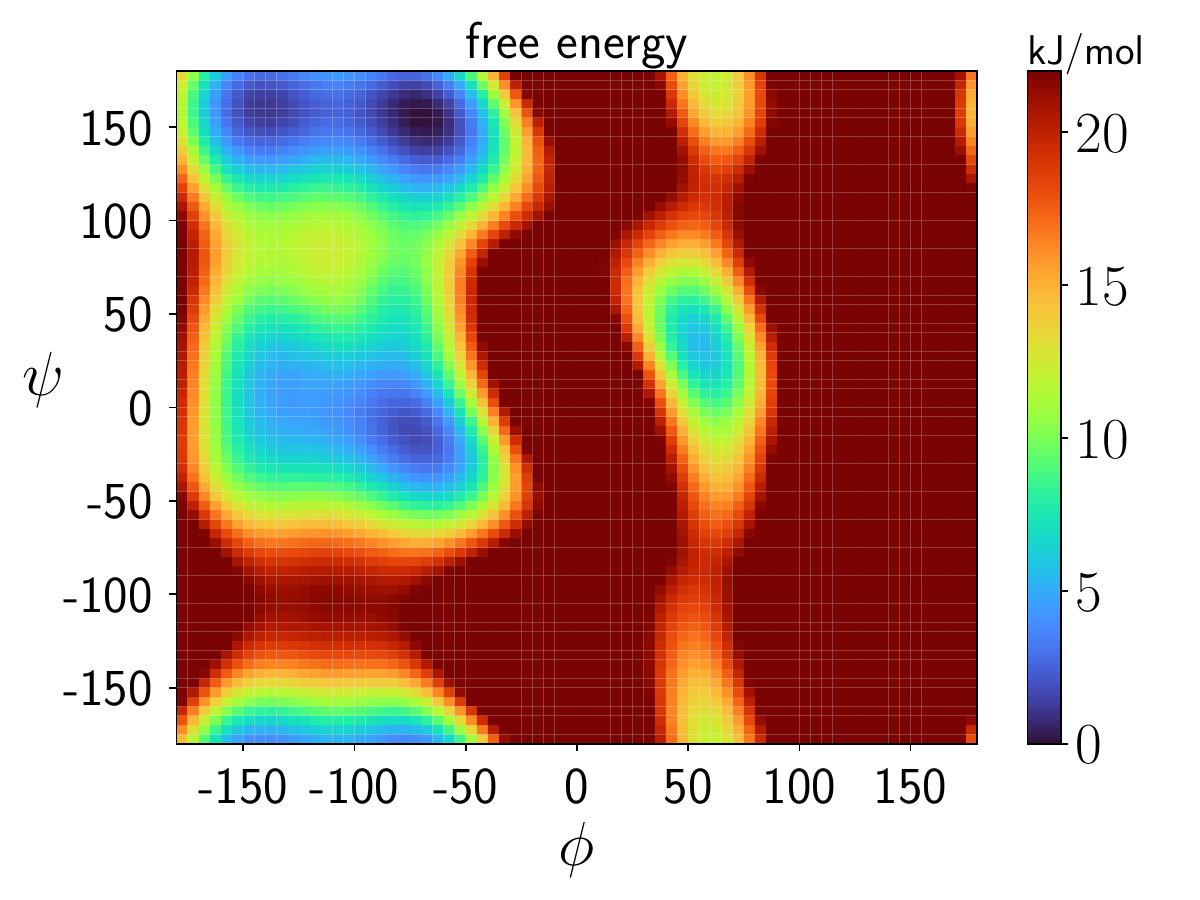}
    \caption{\label{fig:pmf_abf_cv2d}}
    \end{subfigure}
    \begin{subfigure}[t]{0.40\linewidth}
    \centering
    \includegraphics[width=0.95\textwidth]{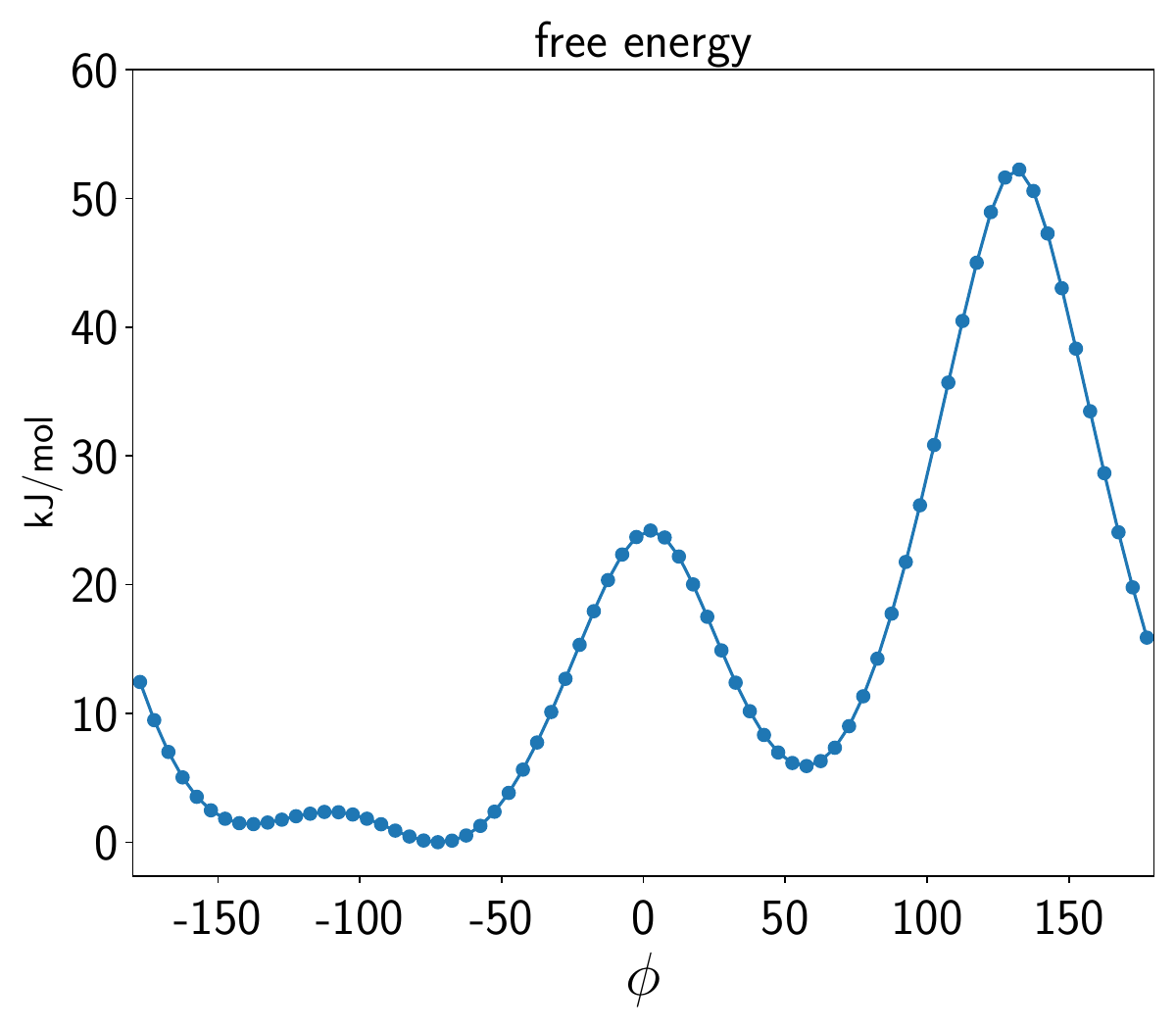}
    \caption{\label{fig:pmf_abf_cv1d}}
\end{subfigure}
\caption{(a) Free energy profile associated to the two dihedral angles $(\phi,\psi)$.  
(b) Free energy profile associated to the dihedral angle $\phi$. }
\end{figure}

In addition to the unbiased trajectory, we also generated two biased
trajectory datasets of the system by performing ABF simulations using the
ColVars module~\cite{Fiorin2013} and the GROMACS package. Specifically, for
the first task (i.e.\ the generation conditioned on both dihedral angles
$\phi$ and $\psi$), we performed an ABF simulation for $t=500\,\textrm{ns}$,
where the CV was defined as the two dihedral angles $\phi$ and $\psi$. The
states were stored every $10$\,ps  and the resulting trajectory dataset
consists of $5\times 10^{4}$ states. For the second task (i.e.\ the generation
conditioned on the dihedral angle $\phi$ alone), we ran an ABF simulation for
$t=100\,\textrm{ns}$, where the CV was defined as the backbone dihedral angle
$\phi$ alone. The states were stored every $1$\,ps, resulting in a trajectory
dataset of $ 10^{5}$ states. Figures~\ref{fig:phi_psi_density_abf_cv2d} and
\ref{fig:phi_psi_density_abf_cv1d} show the empirical densities of the two
trajectory datasets projected onto the space of the two dihedral angles,
respectively. It can be observed that different values of the corresponding CVs are more uniformly sampled under the ABF simulations.
 Figures~\ref{fig:pmf_abf_cv2d} and~\ref{fig:pmf_abf_cv1d} display the free energy profiles associated to the two corresponding CVs (reconstructed using the ColVars module), but this information on free energies is not used in the study below.  
 Finally, coordinates of states in both datasets were preprocessed in the same manner as the unbiased data—centered, aligned to a reference structure, flattened, and then standardized.

For the modeling of the vector field $v$ in~\eqref{fm-ode-z}, we implemented a fully connected feedforward neural network that takes as input the atomic coordinates $x\in \mathbb{R}^{30}$, the temporal parameter $t \in [0,1]$, and a value $z$ that corresponds to the conditioning (either angles or their trigonometric functions). Specifically, in the first task, the value $z$ is defined either as the standardized values of $(\phi, \psi)$ in $\mathbb{R}^2$, or as $z=(\cos\phi,\sin\phi,\cos\psi,\sin\psi)\in\mathbb{R}^4$.  In the second task, $z$ is defined either as the standardized value of $\phi$ in $\mathbb{R}$, or as $z=(\cos\phi,\sin\phi)\in\mathbb{R}^2$. The advantage of using trigonometric functions is that the vector field is guaranteed to be periodic as a function of angles.
In all cases, the feedforward network consists of five hidden layers with 512 hidden units and \texttt{Tanh} activations, while the output layer returns the vector field in $\mathbb{R}^{30}$.

The training of the vector field $v$ was performed by minimizing the objective
\eqref{fm-loss-xi-unbiased-empirical} for 1000 epochs, using the Adam
optimizer with a learning rate of $10^{-3}$, a batch size of 512, and default
PyTorch parameters (e.g.\ momentum coefficients, numerical stability
parameter, and weight decay). For each of the two tasks, the training was carried out both
on the unbiased dataset and on the ABF-biased dataset.  After training, new samples in $\mathbb{R}^{30}$ can be generated  conditioned on the target CV value (i.e.\ the target value of $(\phi, \psi)$ in the first task, or the target value of $\phi$ in the second task) by integrating the ODE~\eqref{fm-ode-z} from random initial positions sampled from the standard Gaussian distribution. The generated samples are then denormalized and reshaped to obtain three-dimensional coordinates of the $10$ non-Hydrogen atoms. 

\begin{figure}[t!]
    \centering
    \begin{subfigure}[t]{0.45\linewidth}
        \centering
        \includegraphics[width=\linewidth]{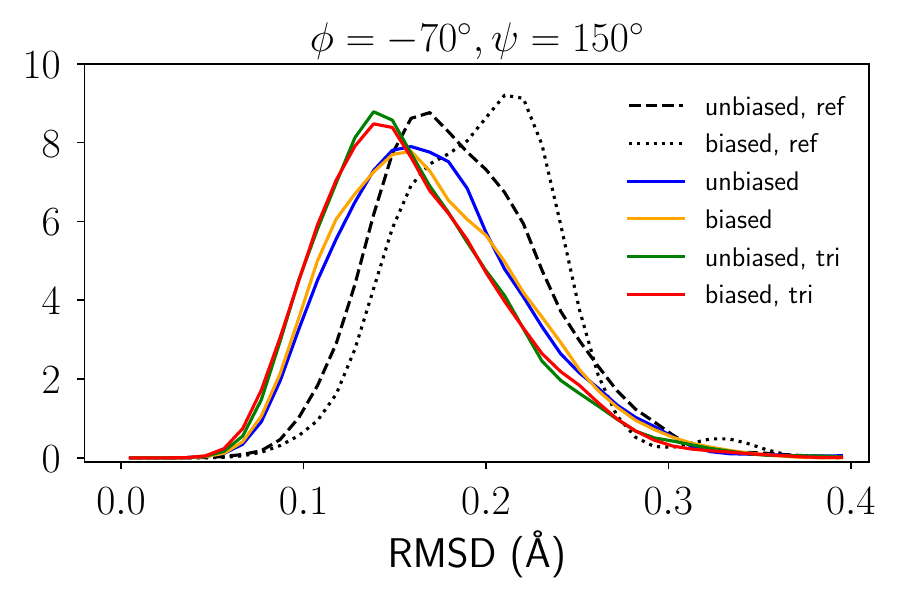}
        \caption{}
        \label{fig:kde_rmsd_0}
    \end{subfigure}
    \begin{subfigure}[t]{0.45\linewidth}
        \centering
        \includegraphics[width=\linewidth]{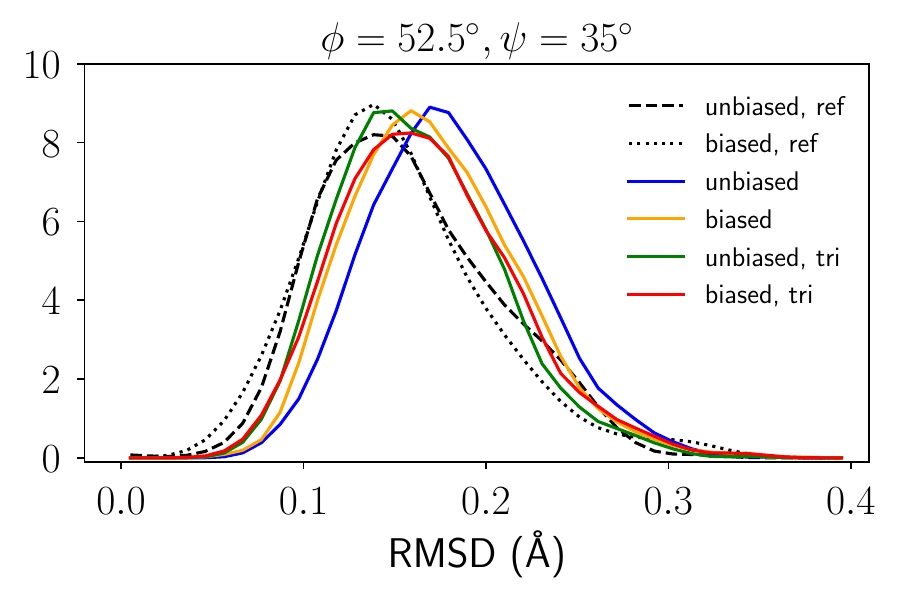}
        \caption{}
        \label{fig:kde_rmsd_1}
    \end{subfigure}
    \caption{Density estimations of RMSD values (in\ \AA) for two target dihedral angles: (a) $(\phi, \psi) = (-70^\circ, 150^\circ)$ and (b) $(\phi, \psi) = (52.5^\circ, 35^\circ)$ in the first task. In both figures, the line ``unbiased, ref'' (resp.\ ``biased, ref'') corresponds to the density of RMSDs computed for the configurations in the unbiased (resp.\ ABF-biased) trajectory dataset whose dihedral angles are close to the target dihedral values (see the text for details). The line with label ``unbiased'' (resp.\ ``biased'')
     corresponds to the model trained on the unbiased (resp.\ ABF-biased) dataset, where the
     standardized angles are directly passed to the neural network. The line
     with label ``unbiased, tri'' (resp.\ ``biased, tri'') corresponds to the
     model trained on the unbiased (resp.\ ABF-biased) dataset, where the values $(\cos\phi,\sin\phi,\cos\psi,\sin\psi)$ are passed to the neural network.  
     } 
    \label{fig:cv2d_rmsd_kde}
\end{figure}

\paragraph{Results for the first task.}
To assess the quality of the generated configurations, we chose the target dihedral angles  
$(\phi, \psi)\in \{ (-70^\circ, 150^\circ), (52.5^\circ, 35^\circ)\}$, which
correspond to a state in the densely populated region and a state in the
secondary metastable region, respectively (see
Figure~\ref{fig:phi_psi_density_unbiased}). For each of these values $(\phi, \psi)$, a configuration in the unbiased trajectory whose dihedral angles are
closest to the target value $(\phi, \psi)$ was identified as a reference, and
2000 samples were generated conditioned on the target angles for each of the
trained models (i.e.\ models trained on the unbiased dataset and on the ABF-biased dataset, with conditioning $z$ being either angles or trigonometric functions). The root mean square deviation (RMSD) values of the generated configurations from the reference (after Kabsch alignment) were computed. Figures~\ref{fig:kde_rmsd_0}--\ref{fig:kde_rmsd_1} show the estimated densities of the RMSDs. Moreover, we selected the configurations in the dataset
whose dihedral values belong to the same cell as the target dihedral values
$(\phi, \psi)$, when binned into the uniform grid of $[-180^\circ,
180^\circ]\times [-180^\circ, 180^\circ]$ with width $10^\circ$. The RMSD
values of the selected configurations from the reference were computed and the
estimated densities are also shown in Figures~\ref{fig:kde_rmsd_0}--\ref{fig:kde_rmsd_1} for comparison. It can be observed that the density profiles of RMSDs computed for the generated configurations are similar to the density profiles computed for the configurations selected from the trajectory datasets, indicating that the trained models are able to produce consistent configurations conditioned on the target dihedral angles.
The fact that the density profiles are unimodal is also consistent with the fact that the configuration of alanine dipeptide is largely determined by its two dihedral angles.

\begin{figure}[ht!]
    \centering
    \begin{subfigure}[t]{0.35\textwidth}
        \centering
        \includegraphics[width=\textwidth]{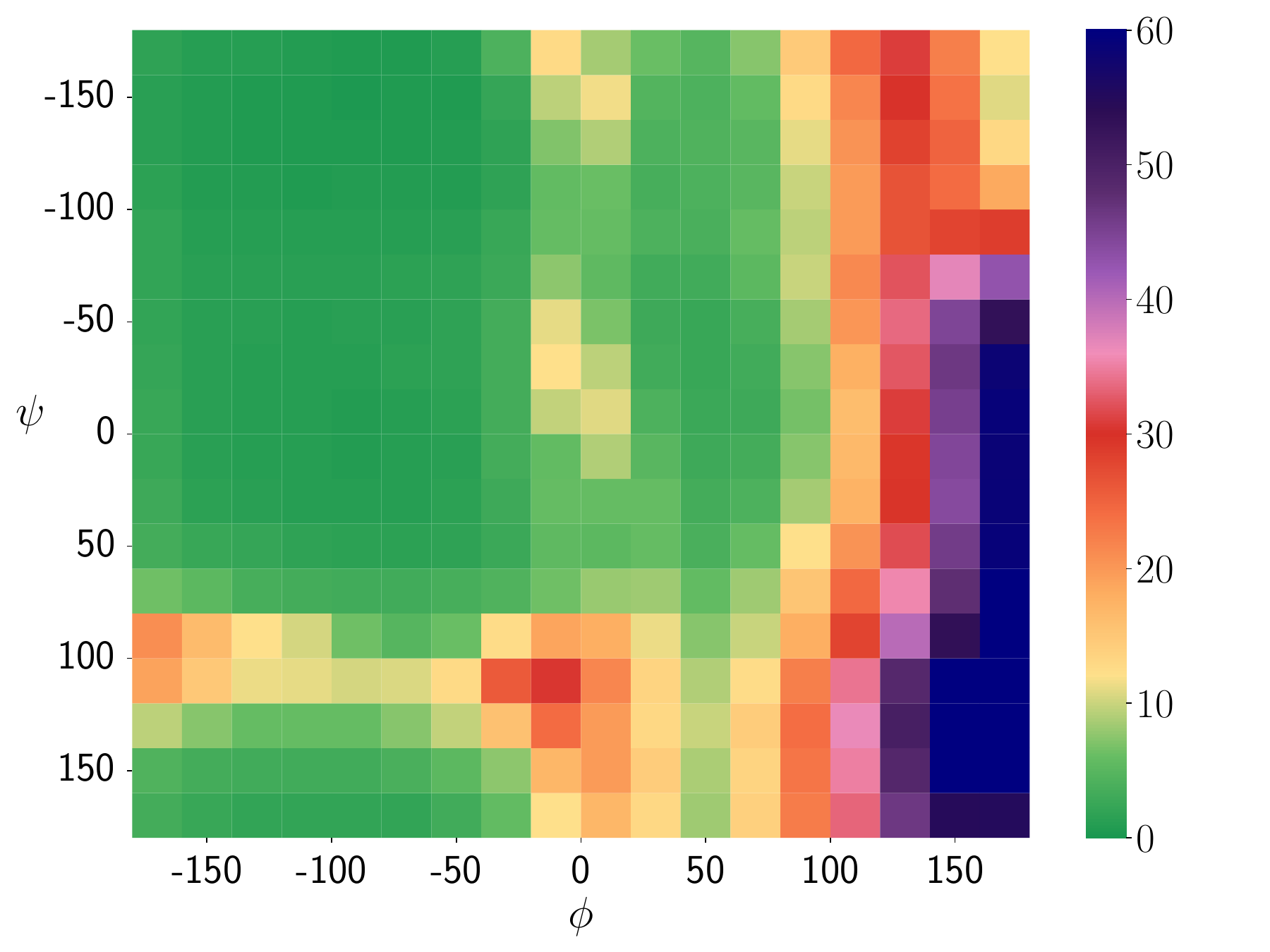}
        \caption{}
        \label{fig:deviation_unbiased}
    \end{subfigure}
    \hspace{0.02\textwidth}
    \begin{subfigure}[t]{0.35\textwidth}
        \centering
        \includegraphics[width=\textwidth]{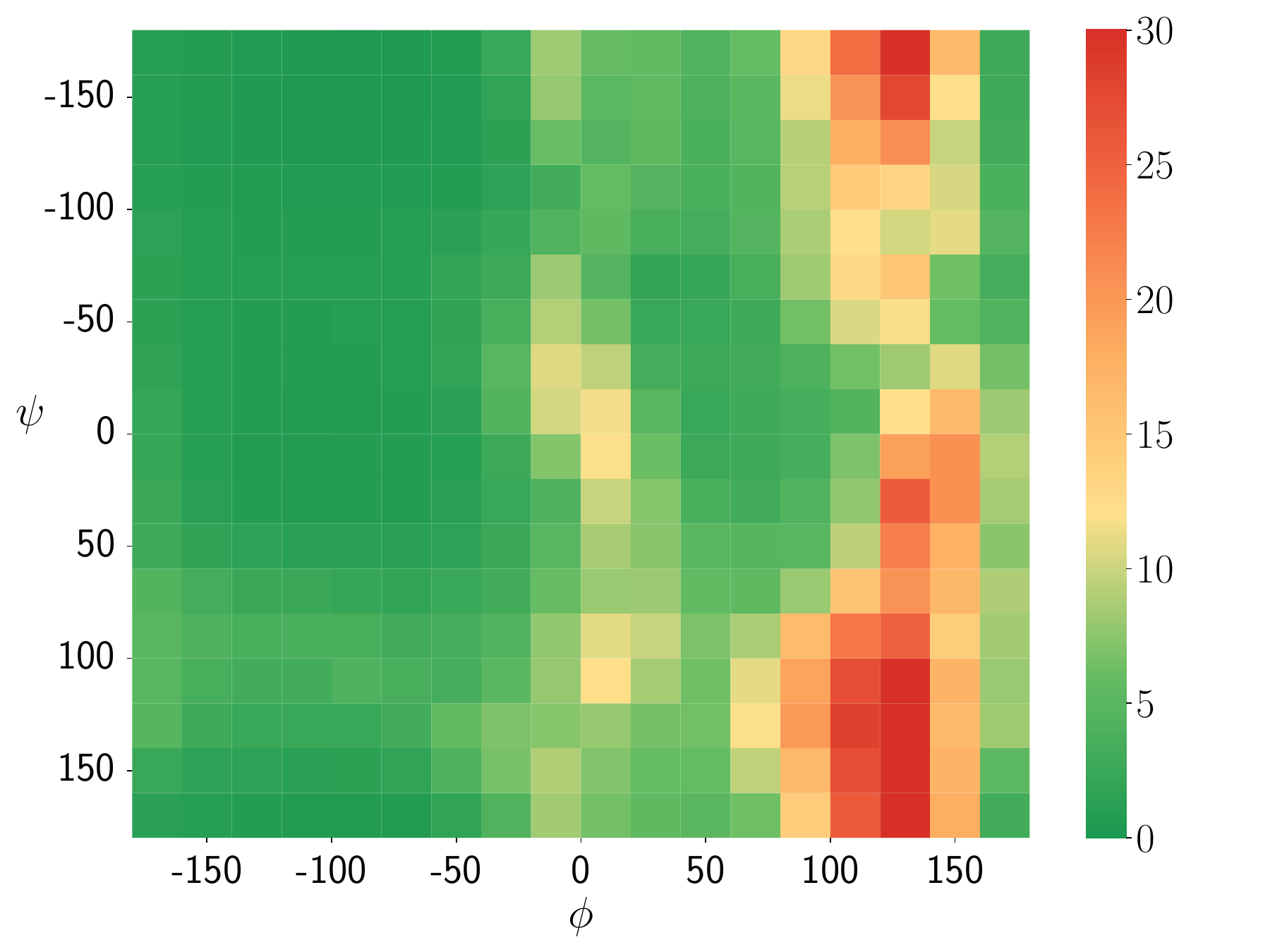}
        \caption{}
        \label{fig:deviation_unbiased_tri}
    \end{subfigure}
       \begin{subfigure}[t]{0.35\textwidth}
        \centering
        \includegraphics[width=\textwidth]{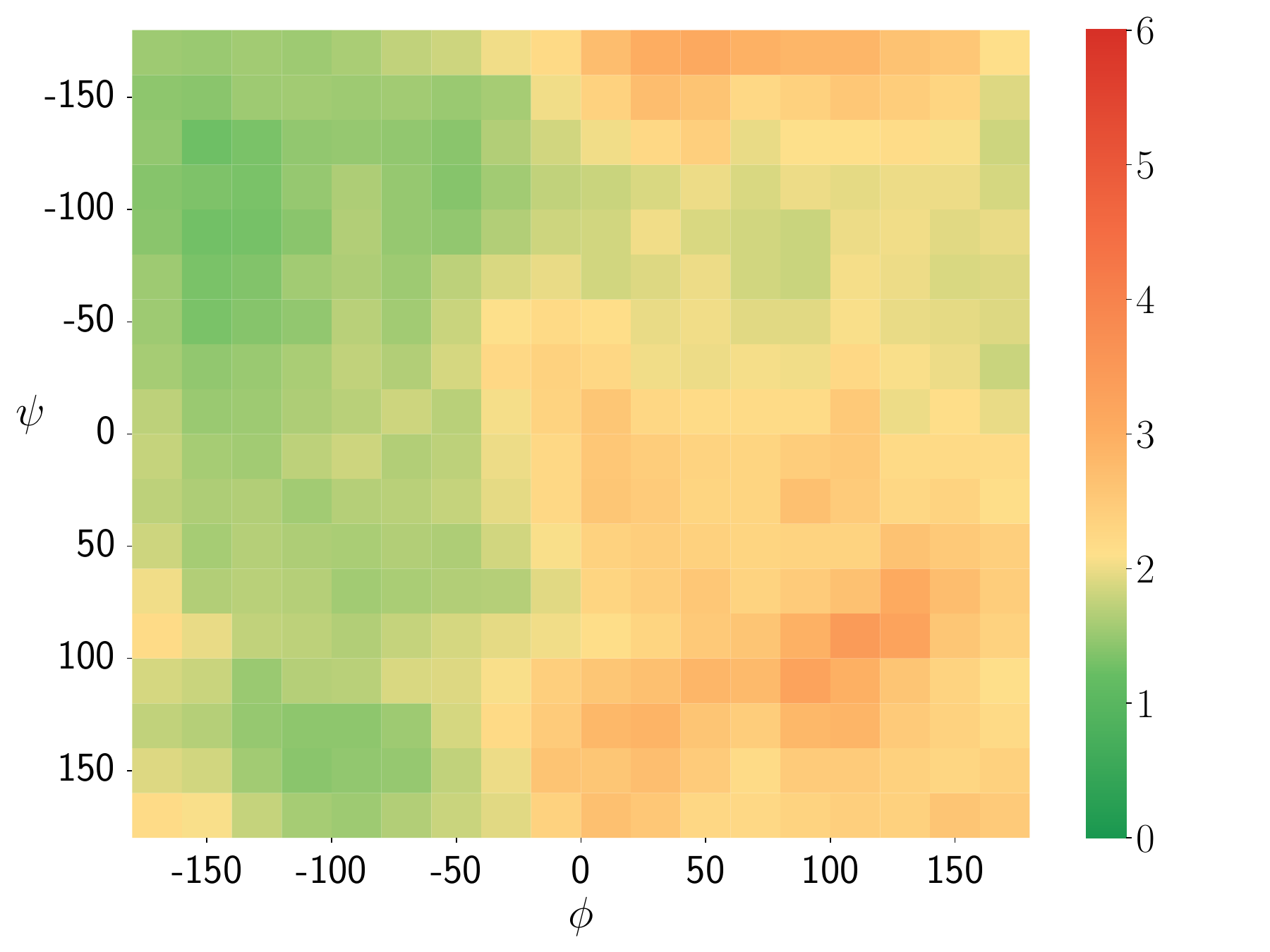}
        \caption{}
        \label{fig:deviation_biased}
    \end{subfigure}
    \hspace{0.02\textwidth}
    \begin{subfigure}[t]{0.35\textwidth}
        \centering
        \includegraphics[width=\textwidth]{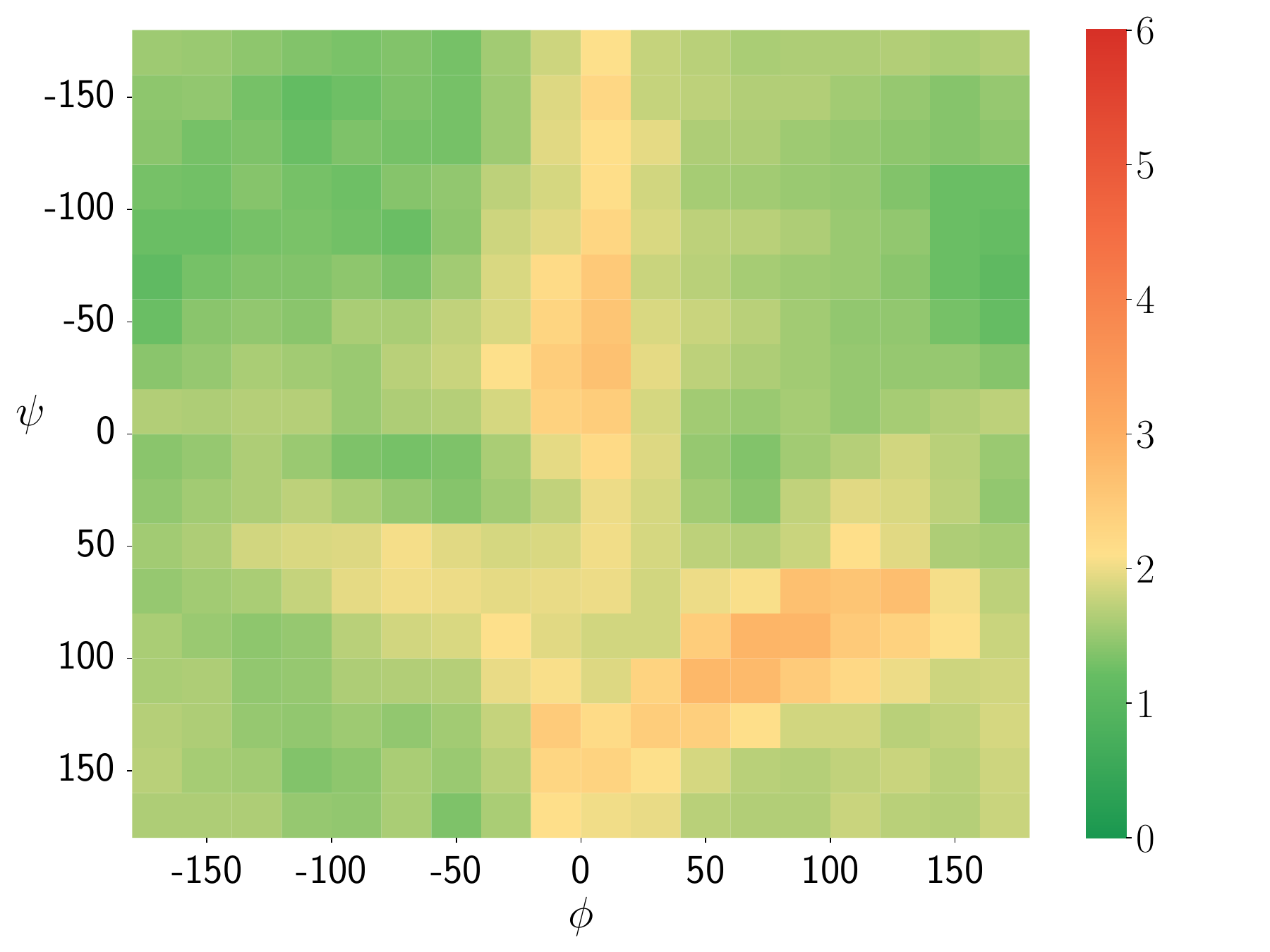}
        \caption{}
        \label{fig:deviation_biased_tri}
    \end{subfigure}
    \caption{Heat maps of the mean deviation (i.e.\ the mean circular distance
    in degrees) between the target CV value $(\phi, \psi)$ and the dihedral
    angles of generated configurations in the first task. (a) Model trained on the unbiased trajectory data, where the standardized dihedral angles are used as input of the neural network. 
    (b) Model trained on the unbiased trajectory data, where the trigonometric functions $(\cos\phi,\sin\phi,\cos\psi,\sin\psi)\in\mathbb{R}^4$ are used as input of the neural network. (c) Model trained on the ABF-biased trajectory data, where the standardized dihedral angles are used as input of the neural network. (d) Model trained on the ABF-biased trajectory data, where the trigonometric functions $(\cos\phi,\sin\phi,\cos\psi,\sin\psi)\in\mathbb{R}^4$ are used as input of the neural network. Lower deviation indicates higher fidelity of the generated ensemble to the target CVs.}
    \label{fig:deviation_heatmaps}
\end{figure}

To assess the angular consistency of the generated configurations across the
dihedral angle space, we computed heat maps of mean deviations (see
\eqref{def-deviation-z-paragraph}) per $(\phi, \psi)$-bin for the trained
models, as shown in Figure~\ref{fig:deviation_heatmaps}.  For this analysis,
the $\phi$ and $\psi$ angles in the domain $[-180^\circ, 180^\circ]\times
[-180^\circ, 180^\circ]$ were binned using a uniform grid of equal width
$20^\circ$ (in both directions), and the centers of grid cells were used as
target CV values.  For each target CV value, $N=200$ configurations were
generated using the trained ODE models. Figure~\ref{fig:deviation_unbiased} and Figure~\ref{fig:deviation_unbiased_tri}
show the resulting mean deviations for the models trained on the unbiased trajectory data, where the standardized dihedral angles 
and the trigonometric functions $(\cos\phi,\sin\phi,\cos\psi,\sin\psi)\in\mathbb{R}^4$ were passed to the neural network, respectively.
  For both models, small deviations are concentrated in the high-probability
  regions of the CV space. Comparing these two figures, it is clear that using
  trigonometric functions is advantageous, as it leads to lower mean
  deviations in the region close to the right boundary
  $\phi=180^\circ$ in the CV space, since the data points close to the left boundary
  $\phi=-180^\circ$ can contribute to the learning in this region thanks to
  the periodicity of the vector field (see
  Figure~\ref{fig:phi_psi_density_unbiased}).  Similarly,
  Figures~\ref{fig:deviation_biased} and \ref{fig:deviation_biased_tri} show
  the mean deviations for the models trained on the ABF-biased trajectory
  data, where the standardized dihedral angles and their trigonometric
  functions were used as inputs of the neural network, respectively. Comparing
  to the models trained on the unbiased trajectory data, lower deviations are observed across the CV space for the models trained on the ABF-biased trajectory data. Since in this case the CV space is well covered by the ABF-biased trajectory data, only slight improvement is achieved when trigonometric functions are used (Figure~\ref{fig:deviation_biased_tri}). 

\paragraph{Results for the second task.}
Regarding the generative modeling task conditioned on the angle $\phi$ alone,
we tested the models trained on the unbiased trajectory data (Figure~\ref{fig:phi_psi_density_unbiased}) and on the ABF-biased data (Figure~\ref{fig:phi_psi_density_abf_cv1d}), and for each case the vector field $v$ takes either the standardized angle $\phi$ or the trigonometric functions $(\cos\phi,\sin\phi)\in\mathbb{R}^2$ as inputs for the conditioning. 

\begin{figure}[t!]
    \centering
    \begin{subfigure}[t]{0.313\textwidth}
        \centering
        \includegraphics[width=\textwidth]{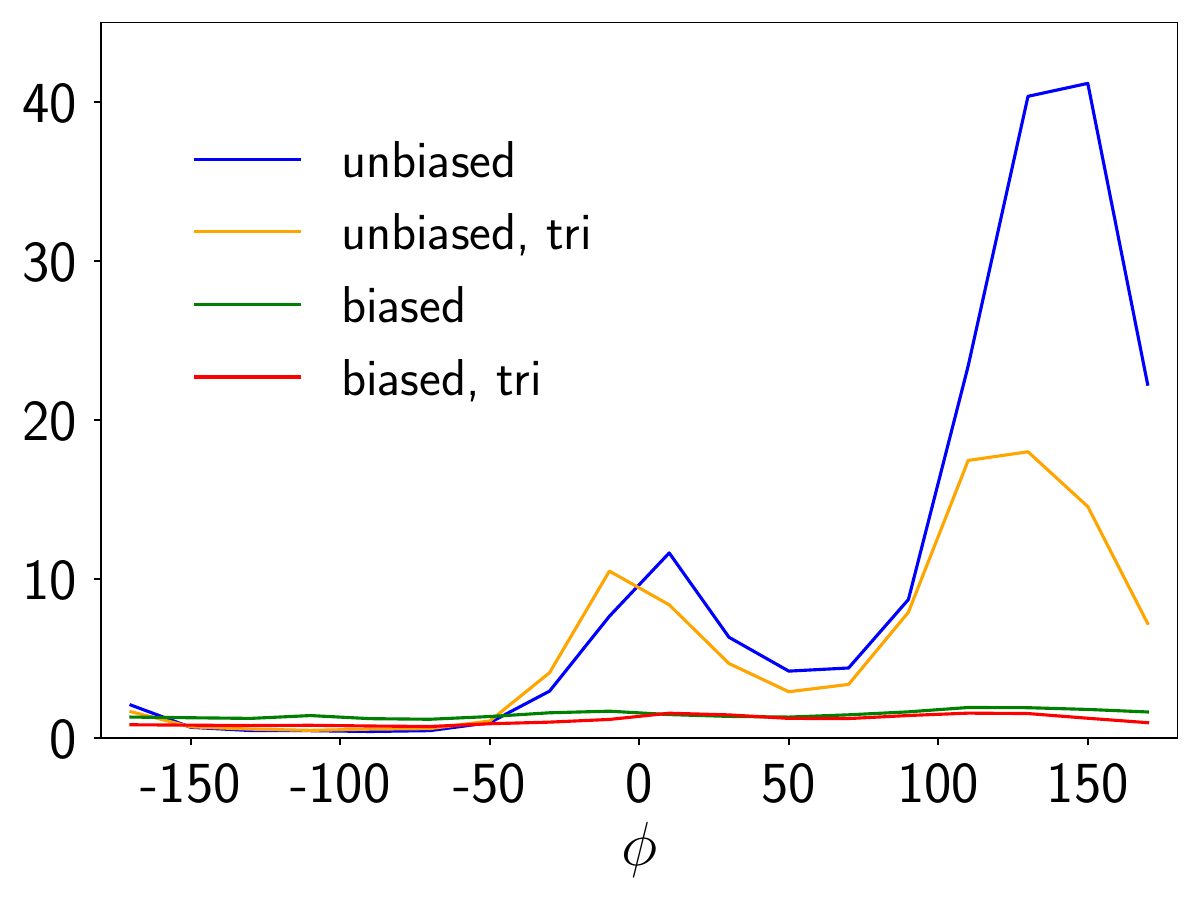}
        \caption{}
        \label{fig:cv1d_mean_dev_cmp}
    \end{subfigure}
    \begin{subfigure}[t]{0.33\textwidth}
        \centering
        \includegraphics[width=\textwidth]{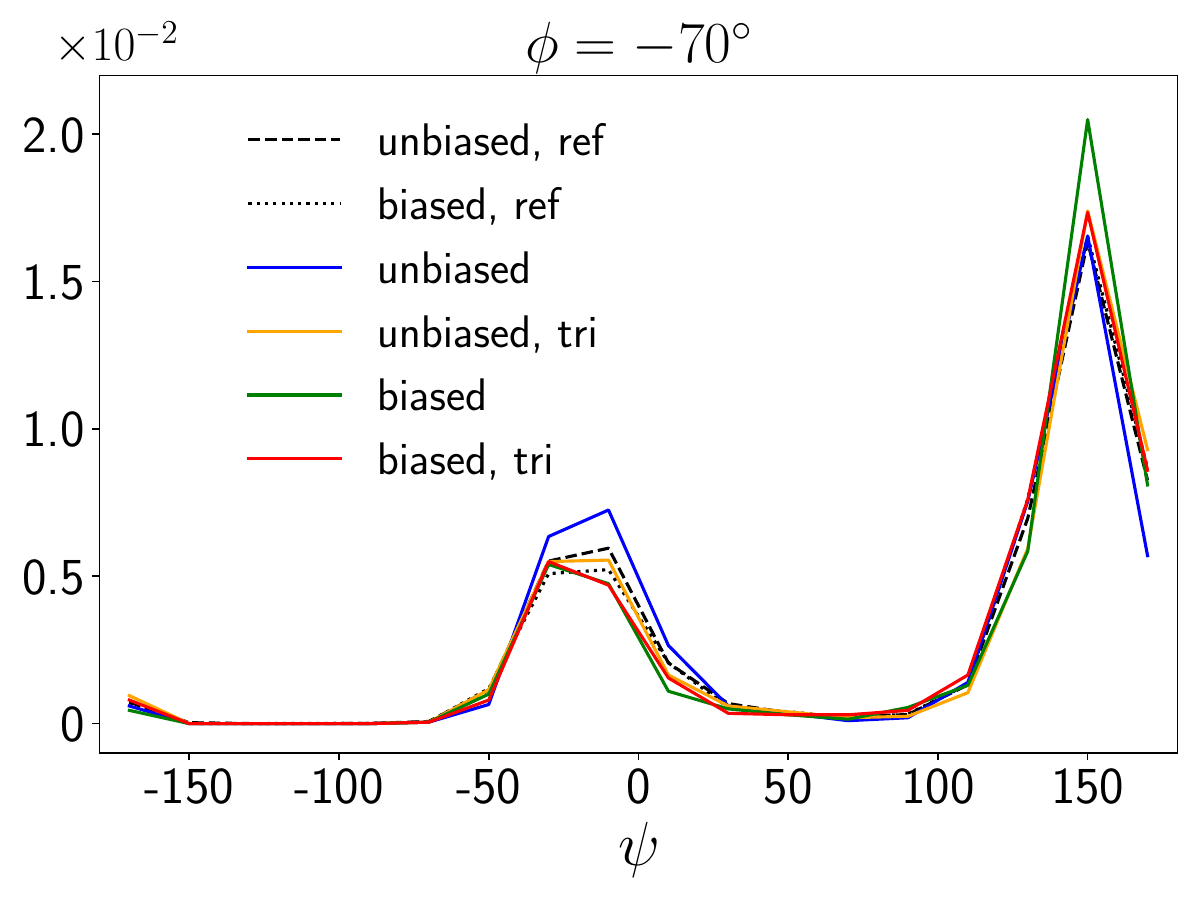}
        \caption{}
        \label{fig:cv1d_density_cmp_0}
    \end{subfigure}
    \begin{subfigure}[t]{0.33\textwidth}
        \centering
        \includegraphics[width=\textwidth]{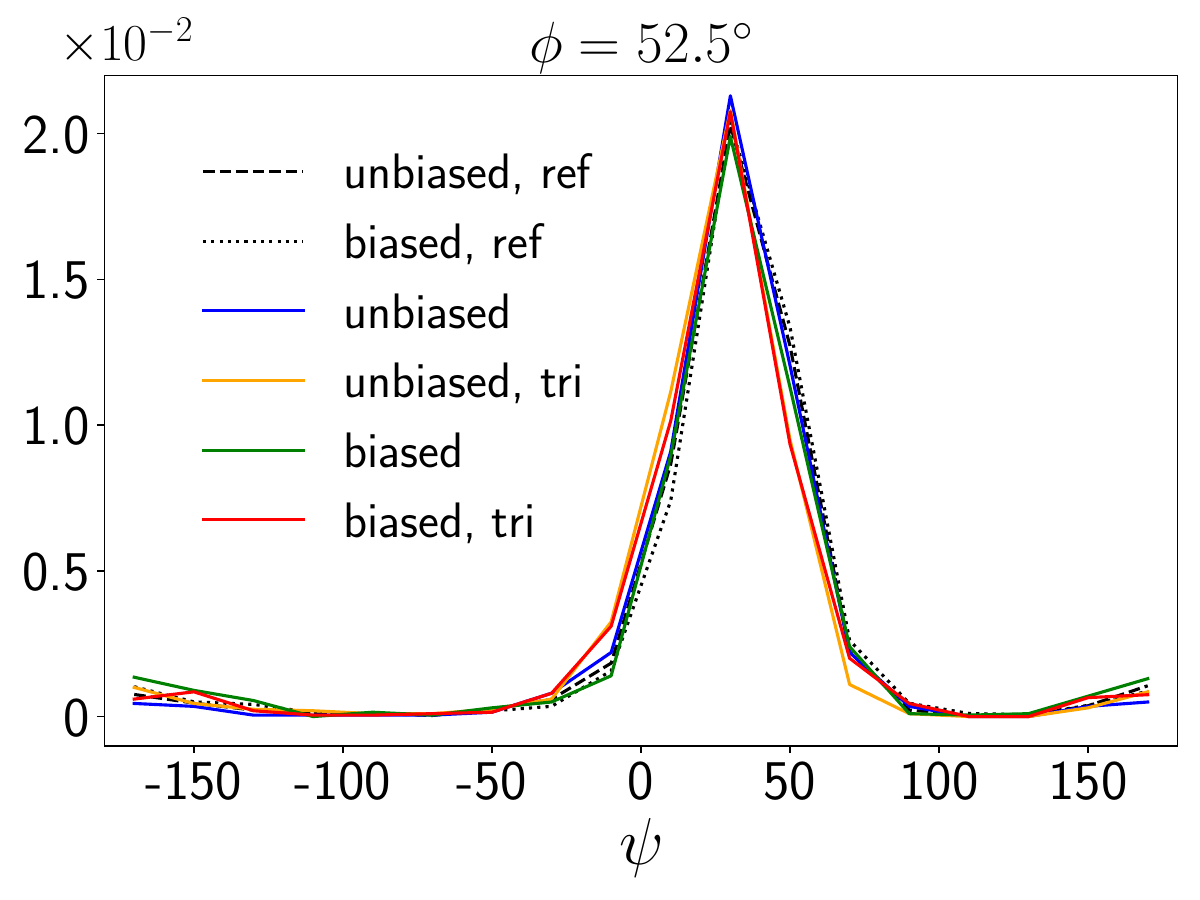}
        \caption{}
        \label{fig:cv1d_density_cmp_1}
    \end{subfigure}
    \caption{Results for the second task. (a) Mean deviation (in degrees) as a function of the target value $\phi$. (2) Estimated densities of $\psi$ conditioned on $\phi=-70^\circ$. 
    (3) Estimated densities of $\psi$ conditioned on $\phi=52.5^\circ$. 
    In each figure, the line with label ``unbiased'' (resp.\ ``biased'') corresponds to the
    model trained on the unbiased (resp.\ ABF-biased) trajectory dataset,
    where the model takes the standardized value of $\phi$ as an input. The
    line with label ``unbiased, tri'' (resp.\ ``biased, tri'')  corresponds to
    the model trained on the unbiased (resp.\ ABF-biased) trajectory dataset,
    where the model takes the trigonometric functions
    $(\cos\phi,\sin\phi)\in\mathbb{R}^2$ as input. In (b) and (c), the line
    with labels ``unbiased, ref'' (resp.\ ``biased, ref'') corresponds to the
    density of $\psi$ estimated using configurations in the unbiased (resp.\ ABF-biased) trajectory dataset.}
    \label{fig:cv1d}
\end{figure}

The interval $[-180^\circ, 180^\circ]$ was divided into subintervals of equal length $20^\circ$ and the centers of the subintervals were chosen as target CV values of $\phi$. For each of these target values and for each of the trained models, $200$ samples were generated, based on which the mean deviations (see \eqref{def-deviation-z-paragraph}) were computed. The results are shown in Figure~\ref{fig:cv1d_mean_dev_cmp}. Similar to the results in the first task, when the unbiased trajectory data is used in training, low deviations are achieved for values of $\phi$ that are frequently taken by the configurations in the dataset, whereas the deviations are high for values of $\phi$ that are rarely taken by configurations in the dataset.  In this case, the use of trigonometric functions $(\cos\phi,\sin\phi)\in\mathbb{R}^2$ helps to reduce the deviations for target $\phi$ values that are close to $180^\circ$. When the ABF-biased trajectory data is used in training, the model accuracy is greatly improved, as the deviations are low across different target values $\phi$. In this case, the advantage of using trigonometric functions is less obvious since the entire space of $\phi$ is well covered by the data (Figure~\ref{fig:phi_psi_density_abf_cv1d}). 

To further assess the quality of the models, we studied the distribution of
$\psi$ conditioned on the target values $\phi=-70^\circ$ and
$\phi=52.5^\circ$. For each of these target values and for each of the trained
models, 1000 samples were generated, based on which the empirical density of
the angle $\psi$ was estimated. For comparison, the density of $\psi$
conditioned on the target value $\phi$ was also estimated using configurations
in the unbiased trajectory dataset and configurations in the ABF-biased
trajectory dataset (by selecting configurations whose angle $\phi$ belongs to
the same cell as the target value of $\phi$ when binned using a uniform grid
with width $20^\circ$). The results are shown in
Figures~\ref{fig:cv1d_density_cmp_0} and~\ref{fig:cv1d_density_cmp_1}. As can
be observed there, for each of the trained models, the distribution of the angle $\psi$ of the generated samples matches well with the conditional distributions estimated using configurations in the trajectory datasets. 

Finally, as in the previous example, the projection method introduced in
Section~\ref{subsec-projection} can be applied as a post-processing step to
modify the generated configurations so that they satisfy the constraints on
the target dihedral angles exactly. We refer to the previous work~\cite[Appendices~B.2 and~B.4]{rddpm} for a related numerical study. 

\FloatBarrier 

\section{Conclusion}
\label{sec-conclusion}
In this work, we have introduced a general framework for the generative modeling of conditional distributions on level-sets of CVs using flow-matching neural ODEs. By conditioning the learned flows on CV values, the model can generate samples across different level-sets without requiring explicit geometric information. To enhance the generation accuracy on level-sets in low-probability regions, we proposed to leverage biased trajectory data from enhanced sampling techniques such as ABF. Validation on concrete datasets demonstrates that our approach accurately reproduces conditional distributions and successfully exploits biased data to improve learning in scarcely sampled regions. 

This study opens several promising directions for future work. First,  it is interesting to apply the method to more complex molecular systems and study how the method performs (in terms of the amount of required data and the model quality) when the size of the molecular system increases, especially in the case where the target conditional distributions are multimodal. 
For such applications, it could be the case that more advanced architectures (e.g. using featurization or transformers) would improve the generative capabilities. Second, a natural extension would be to combine the approach proposed in this work with methods for identifying good CV maps and methods for propagating the effective (reduced) dynamics in the CV space~\cite{effective_dynamics,effective_dyn_2017}. Such a combination would lead to generative modeling approaches that can generate new trajectories of high-dimensional complex stochastic dynamics while taking the essential dynamics into account.   

%Finally, it is clear that using some featurization could positively impact the performance of the method (instead of relying on Cartesian coordinates, passing to internal variables, or using descriptors akin to the ones used for machine learned force fields~\cite{MLFF_review}).
%Let us also emphasize that the architectures considered in this work are quite simple. For more complex systems, the conditional measures could be multimodal, which makes the generative approach more challenging. One may for instance think of starting from a mixture of Gaussians instead of a single Gaussian, similarly to what is done in GMVAEs (Gaussian Mixture Variational AutoEncoders)~\cite{GMVAE}. 

Overall, our study provides a general and practical approach for conditional generative modeling on manifolds defined by CVs, offering both methodological flexibility and the potential for applications in molecular modeling and beyond.

\section{Acknowledgements}
Fatima-Zahrae Akhyar acknowledges Zuse institute Berlin for hosting her internship during which this research was conducted. Wei Zhang has been funded through DFG’s ``Eigene Stelle'' program (project 524086759) and has been supported by the Young Investigator Program of the DFG-grant CRC 1114 ``Scaling Cascades in Complex Systems'' Project Number 235221301. 
Gabriel Stoltz was funded by the European Research Council (ERC) under the European Union's Horizon 2020 research and innovation programme (project EMC2, grant agreement No 810367), and was supported by Hi! PARIS and ANR/France 2030 program (ANR-23-IACL-0005). Christof Sch\"{u}tte has been supported 
through DFG-grant CRC 1114 
(Project A04 ``Efficient calculation of slow and stationary scales in molecular dynamics'', Project A05 ``Probing scales in equilibrated systems by optimal nonequilibrium forcing'',  Project B03 ``Multilevel coarse graining of multiscale problems'', and Project C03 ``Multiscale modelling and simulation for spatiotemporal master equation'').

\bibliographystyle{abbrv}
\bibliography{references}
\end{document}